%% file: Main.tex
\documentclass[10pt,journal,compsoc]{IEEEtran} 

\usepackage{bm}
\usepackage{amsmath}
\usepackage{epsf}
\usepackage{graphics}
\usepackage{ amssymb }
\usepackage[dvips]{graphicx}
\usepackage{mathtools}
\usepackage{epsfig}
\usepackage{cite}
\usepackage[linesnumbered,ruled,vlined]{algorithm2e}
\usepackage{colortbl}
\usepackage{color}
\usepackage{hyperref} 
\usepackage{enumitem}
\usepackage{soul,xcolor}

\usepackage{bm}
\usepackage{amsmath}
\usepackage{epsf}
\usepackage{graphics}
\usepackage{ amssymb }
\usepackage[dvips]{graphicx}
\usepackage{epsfig}
\usepackage{cite}
\usepackage[linesnumbered,ruled,vlined]{algorithm2e}
\usepackage{graphicx}
\usepackage{epsfig}
\usepackage{latexsym}
\usepackage{amsfonts}
\usepackage{here}
\usepackage{rawfonts}
\usepackage[utf8]{inputenc}
\usepackage[english]{babel}
\usepackage{amsmath}
\usepackage{amsfonts}
\usepackage{amssymb}
\usepackage{color}
\usepackage{bm}
\usepackage{listings}
\usepackage{caption}
\usepackage{multicol}
\usepackage{tabularx,booktabs}
\newcolumntype{Y}{>{\centering\arraybackslash}X}
\makeatletter
\newcommand\notsotiny{\@setfontsize\notsotiny{6.31415}{7.1828}}
\makeatother
\usepackage{amssymb}
\usepackage{amsthm}
\usepackage{graphicx}
\usepackage{epstopdf}
\usepackage{listings}
\usepackage{float}
\usepackage{amsmath}
\usepackage{amssymb}
\usepackage{amsfonts}
\usepackage{epstopdf}

\usepackage{multirow}
\usepackage{amscd}
\usepackage{mathrsfs}
\usepackage{graphicx}
\usepackage{color}
\usepackage{url}
\usepackage{bm}
\usepackage{setspace}
\usepackage{footnote}
\usepackage{xcolor}
\definecolor{lightblue}{RGB}{173, 216, 230}  
\newtheorem{remark}{Remark}

\usepackage[]{algorithm2e}

\addto\captionsenglish{}
\usepackage{bbm}

\usepackage{multicol}
\usepackage{mathtools}

\usepackage{lipsum,graphicx,subcaption}

\usepackage{diagbox}
\usepackage{hyperref}
\usepackage[protrusion=true,expansion=true]{microtype}
\pdfoutput=1
\usepackage[font=footnotesize]{caption}
\allowdisplaybreaks[4]

\usepackage{graphicx}
\usepackage{grffile}
\usepackage{tabularx}
\newcounter{term}[section]

\renewcommand\theterm{\alph{term}}
\makeatletter
\newcommand{\vast}{\bBigg@{4}}

\newcommand{\Vast}{\bBigg@{5}}
\makeatother
\newcommand\semiHuge{\fontsize{22.7}{31.38}\selectfont}
\usepackage{soul,xcolor}

\usepackage[most]{tcolorbox}
\tcbset{colback=yellow!10!white, colframe=black!50!black, 
        highlight math style= {enhanced, 
            colframe=black,colback=red!10!white,boxsep=0pt}
        }
        \allowdisplaybreaks

\begin{document} 

\title{{\semiHuge Two-Timescale Model Caching and Resource Allocation for Edge-Enabled AI-Generated Content Services}}
\author{Zhang Liu,~\IEEEmembership{Student Member,~IEEE}, Hongyang Du,~\IEEEmembership{Member,~IEEE}, Xiangwang Hou,~\IEEEmembership{Student Member,~IEEE},\\ Lianfen Huang,~\IEEEmembership{Member,~IEEE}, Seyyedali Hosseinalipour,~\IEEEmembership{Member,~IEEE}, Dusit Niyato,~\IEEEmembership{Fellow,~IEEE},\\ and Khaled Ben Letaief,~\IEEEmembership{Fellow,~IEEE}
\thanks{\emph{Z. Liu (zhangliu@stu.xmu.edu.cn) and L. Huang (lfhuang@xmu.edu.cn) are with the Department of Informatics and Communication Engineering, Xiamen University, Fujian, China 361102. H. Du (duhy@eee.hku.hk) is with the Department of Electrical and Electronic Engineering, University of Hong Kong, Pok Fu Lam, Hong Kong. X. Hou (xiangwanghou@163.com) is with the Department of Electronic Engineering, Tsinghua University, Beijing, 100084, China. S. Hosseinalipour (alipour@buffalo.edu) is with the Department of Electrical Engineering, University at Buffalo--SUNY, Buffalo, NY 14260. D. Niyato (dniyato@ntu.edu.sg) is with the College of Computing and Data Science, Nanyang Technological University, Singapore 639798. Khaled Ben Letaief (e-mail: eekhaled@ece.ust.hk) is with the Department of Electrical and Computer Engineering, The Hong Kong University of Science and Technology (HKUST), Hong Kong, China. (Corresponding author: Lianfen Huang.)} }
}
\maketitle
\vspace{-9mm}
\setulcolor{red}
\setul{red}{2pt}
\setstcolor{red}

\begin{abstract}
Generative AI (GenAI) has emerged as a transformative technology, enabling customized and personalized AI-generated content (AIGC) services. 
In this paper, we address challenges of edge-enabled AIGC service provisioning, which remain underexplored in the literature. These services require executing GenAI models with billions of parameters, posing significant obstacles to resource-limited wireless edge. We subsequently introduce the formulation of joint model caching and resource allocation for AIGC services to balance a trade-off between AIGC quality and latency metrics. We obtain mathematical relationships of these metrics with the computational resources required by GenAI models via experimentation. Afterward, we decompose the formulation into a model caching subproblem on a long-timescale and a resource allocation subproblem on a short-timescale. Since the variables to be solved are discrete and continuous, respectively, we leverage a double deep Q-network (DDQN) algorithm to solve the former subproblem and propose a diffusion-based deep deterministic policy gradient (D3PG) algorithm to solve the latter. The proposed D3PG algorithm makes an innovative use of diffusion models as the actor network to determine optimal resource allocation decisions. Consequently, we integrate these two learning methods  within the overarching two-timescale deep reinforcement learning (T2DRL) algorithm, the performance of which is studied through comparative numerical simulations.
\end{abstract}
\vspace{-3mm}
\section{Introduction} \label{sec:intro}
\subsection{Background and Overview} \label{subsec:background}
\vspace{-.15mm}
Recent breakthroughs in artificial intelligence (AI) have propelled generative AI (GenAI) into the spotlight, drawing significant attention for its unprecedented ability to automate the creation of a diverse array of AI-generated content (AIGC), including text, audio, and graphics/images~\cite{cao2023comprehensive,qu2024performance}. GenAI aims to produce synthetic data that closely resemble real-world data by learning underlying patterns and characteristics from existing datasets. For example, \emph{ChatGPT}~\cite{brown2020language} generates human-like text based on a given context prompt, while \emph{RePaint}~\cite{lugmayr2022repaint} enables the generation of diverse images from textual descriptions. Building on these advancements, AIGC services have been integrated into various domains, including art, advertising, and education~\cite{anantrasirichai2022artificial}, offering productivity gains and economic growth.

Despite their tremendous potentials, delivering AIGC services relies on the inference process of GenAI models, where the increasing size and complexity of these models present significant challenges for deployment over the wireless edge. For instance, \emph{ChatGPT}, which is built upon GPT-3 with 175 billion parameters, requires 8×48GB A6000 GPUs to perform inference and generate contextually relevant responses to user prompts~\cite{xiao2023smoothquant}. This  can be particularly challenging for the next-generation Internet paradigms, such as Metaverse~\cite{wang2022survey}, which continuously demand high-quality AIGC on personal computers or head-mounted displays. Specifically, the intensive storage and computation demands of GenAI models can limit the accessibility and affordability of provisioning AIGC services.

To address this, cloud data centers with abundant computing and storage capacity can be utilized to train and deploy GenAI models, enabling users to access cloud-based AIGC services through the core network. However, migrating AIGC services to cloud data centers can impose prohibitive traffic congestion on backhaul links~\cite{liu2023rfid, 10304320} and introduce privacy threats within public cloud infrastructure~\cite{xu2024unleashing}. Alternatively, by leveraging wireless edge networks, we can deploy GenAI models closer to users on edge servers co-located with base stations (BSs)~\cite{du2023enabling} to provide lower latency and privacy-aware AIGC services. In this user-edge-cloud continuum, which we aim to investigate, GenAI models are trained and pre-stored in the could, while edge servers are responsible for caching these models and delivering customizable AIGC services to users.

\vspace{-3mm}
\subsection{Motivation and Main Challenges} \label{subsec:challenges}
\begin{figure}[!t]
\includegraphics[width=.48\textwidth]{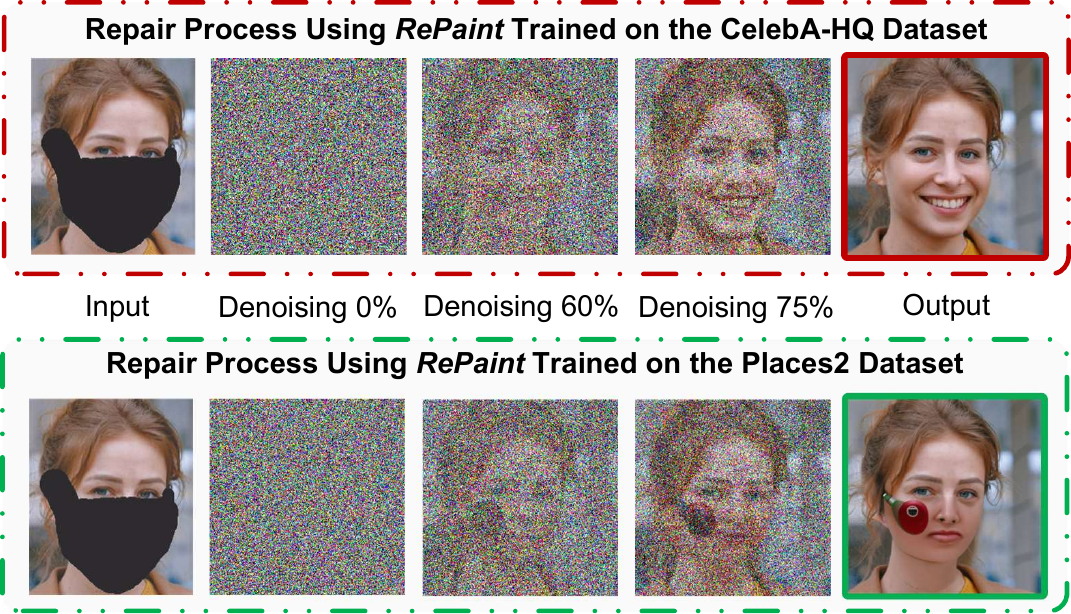}
\centering
\caption{An example of the GenAI model \emph{RePaint}, trained on different datasets, used to repair the same corrupted image.}
\label{fig:dataset_demo}
\end{figure}

\vspace{-.15mm}
While edge servers offer several advantages for delivering AIGC services to users, there are still important issues that need to be addressed. \emph{First, storage-limited edge servers cannot cache all GenAI models simultaneously, leading to performance degradation for users with diverse AIGC service requests.} This degradation occurs because GenAI models rely on massive datasets to learn underlying patterns and generate meaningful outputs, making their performance heavily dependent on the diversity of the training data~\cite{zhang2023text}. To illustrate this, we conducted a study using a corrupted image restoration AIGC service as an example. In this study, the GenAI model \emph{RePaint}, trained separately on the CelebA-HQ dataset (containing images of celebrities' faces) and the Places2 dataset (containing images of various scenes)~\cite{lugmayr2022repaint}, was used to repair a corrupted image. The experiments were conducted on a system equipped with an NVIDIA RTX A5000 GPU. The repair processes shown in Fig.~\ref{fig:dataset_demo} unveil that \emph{RePaint} trained on the human face dataset performed significantly better than the model trained on landscape images. Therefore, due to the diversity of users' AIGC service requests, determining which GenAI models to cache at storage-limited edge servers is  crucial.

\emph{Second, efficiently managing multi-dimensional network resources to deliver low-latency, high-quality AIGC services from edge servers to users is challenging.} Since users may not only send their AIGC requests to edge servers but also retrieve the generated content, both the upload and download processes demand substantial bandwidth. Moreover, GenAI models require significant computational resources to generate content, which must be managed carefully due to the limited resources of edge servers. For instance, the quality of synthetic images produced by diffusion-based GenAI models improves with the number of denoising steps (detailed in Sec.~\ref{subsec:comp_model}). However, the generation latency also increases with the number of denoising steps, necessitating a careful balance of the trade-off between high-quality and low-latency AIGC services, especially in networks with large numbers of users.

\emph{Third, conventional optimization methods for model caching and resource allocation often suffer from high computational complexity, rendering them unsuitable for mobile edge networks, which are subject to temporal variations.} In practical scenarios, user locations and wireless channel conditions vary over time, and the popularity of AIGC services can fluctuate due to dynamic trends. As a result, a solution that is optimal at one point in time may not remain optimal over a longer period. However, existing model caching and resource allocation methods, such as iteration-based algorithms~\cite{9861697},~\cite{9206079} and constraint relaxation-based problem transformations~\cite{abolhassani2022fresh},~\cite{9385953}, either require extensive iterations to converge to a satisfactory solution or suffer from significant performance degradation when the environment undergoes abrupt changes over time. 

\vspace{-3mm}
\subsection{Summary of Contributions} \label{SubSec:Contributions}
\vspace{-.15mm}
\emph{To our knowledge, this is the first study that  optimizes the edge-enabled provisioning of AIGC services by coordinating GenAI model caching and resource allocation decisions in mobile edge networks.} Our main contributions are as follows:
\begin{itemize}[leftmargin=4mm]
\item We formulate the model caching and resource allocation problem in GenAI-enabled wireless networks, which is found to be mixed integer nonlinear programming (MINLP) known to be NP-hard. This makes solving the problem challenging, especially under user mobility, imperfect knowledge of wireless channels, and varying AIGC requests.
\item To tackle the problem, we first divide it into a model caching subproblem on a long-timescale and a resource allocation subproblem on a short-timescale. Since the variables to be solved are discrete and continuous, respectively, we employ a double deep Q-network (DDQN) algorithm to
solve the former subproblem and propose a diffusion-based deep deterministic policy gradient (D3PG) algorithm to address the latter. We integrate these two methods within the overarching two-timescale deep reinforcement learning (T2DRL) algorithm.
\item We conduct experiments using practical GenAI models to develop unified mathematical models that describe the relationships between AIGC quality, service provisioning delay and computational resources. Furthermore, in D3PG algorithm we make an innovative use of diffusion models -- originally designed for image generation -- to determine optimal resource allocation decisions for AIGC provisioning.
\item We validate the effectiveness of our  method through experiments under various simulation settings, demonstrating that T2DRL not only achieves a higher model hitting ratio but also delivers higher-quality, lower-latency AIGC services compared to the  benchmark solutions. This improvement is largely attributed to the generative capabilities of the diffusion model, which enhance action sampling efficiency by progressively reducing noise through the reverse process. 
\end{itemize}


\vspace{-3mm}
\section{Related Works} \label{sec:related_works}
\vspace{-.15mm}
Henceforth, we summarize the contributions of related works and highlight the aspects they have not addressed, which serve as the primary motivations for this work.

\vspace{-3mm}
\subsection{Content Caching in Wireless Networks}\label{subsec:content_caching}
\vspace{-.15mm}
With the rapidly growing demand for diverse multimedia content on platforms like \emph{YouTube}, \emph{Netflix}, and \emph{Facebook}, caching content at the network edge~\cite{9861697,9206079,abolhassani2022fresh,9090328} has emerged as a solution to reduce content provisioning delays and alleviate traffic loads on core networks by avoiding re-transmissions of highly popular contents. Researchers in~\cite{9861697} investigated a joint content caching and task offloading problem, aiming to minimize task completion delays and energy consumption. Researchers in~\cite{9206079} explored the revenue maximization for recommendation-aware, content-oriented wireless caching networks with repair considerations. Researchers in~\cite{abolhassani2022fresh} introduced a caching model for delivering dynamic content to users over the wireless edge, while capturing content freshness through the age-of-version metric. Researchers in~\cite{9090328} focused on caching popular software for mobile edge computing (MEC).

However, unlike traditional content caching, which primarily focuses on storage capacity constraints, GenAI model caching necessitates the careful orchestration of both computing (e.g., denoising steps for creating images) and storage constraints.  More broadly, the unique challenges of edge-enabled AIGC services are yet to be fully explored. Subsequently, the need to evaluate the performance of various GenAI models and develop mathematical models to analyze the performance of diverse AIGC services is a primary motivation behind this work.



\vspace{-3mm}
\subsection{Usage of Deep Reinforcement Learning in Optimization}\label{subsec:DRL_optimization}
\vspace{-.15mm}
Recently, learning-based algorithms, with deep reinforcement learning (DRL) as a notable example~\cite{10076905,10609797,9385375}, have been extensively applied to enhance decision-making and solution design for complex optimization problems. Unlike conventional optimization methods, DRL employs deep neural networks (DNNs) to learn the relationship between a problem's state space (e.g., AICG service popularity) and its action space (e.g., model caching decisions). Researchers in~\cite{10076905} utilized a DDQN algorithm to solve the joint service caching, resource allocation, and computation offloading problem in a cooperative MEC system. Researchers in~\cite{10609797} proposed a deep deterministic policy gradient (DDPG) method to jointly optimize service caching, collaborative offloading, and resource allocation in multi-access edge computing. Researchers in~\cite{9385375} introduced a  DRL algorithm based on the integrated of DDQN and dueling DQN to optimize edge caching and radio resource allocation. 

In this work, we take the fist step towards tailoring DRL techniques for the provisioning of AIGC services at the network edge. However, the uncertainties introduced by user movement, wireless channel fluctuations, and changing AIGC service requests make the DRL state space complex and highly dynamic. This, in turn, renders the use of common multi-layer perceptron (a type of fully connected DNN) in DRL architectures ineffective due to challenges with exploration-exploitation trade-offs and the potential for convergence to suboptimal policies~\cite{10529221}. This emphasizes the need to explore novel learning-based techniques, which we address by introducing diffusion models into the DRL architecture. 


\vspace{-2mm}
\section{System Model} \label{sec:system_model}
\vspace{-1mm}
\subsection{Network Outline}\label{subsec:system_architecture}
\begin{figure}[t!]
\includegraphics[width=.48\textwidth]{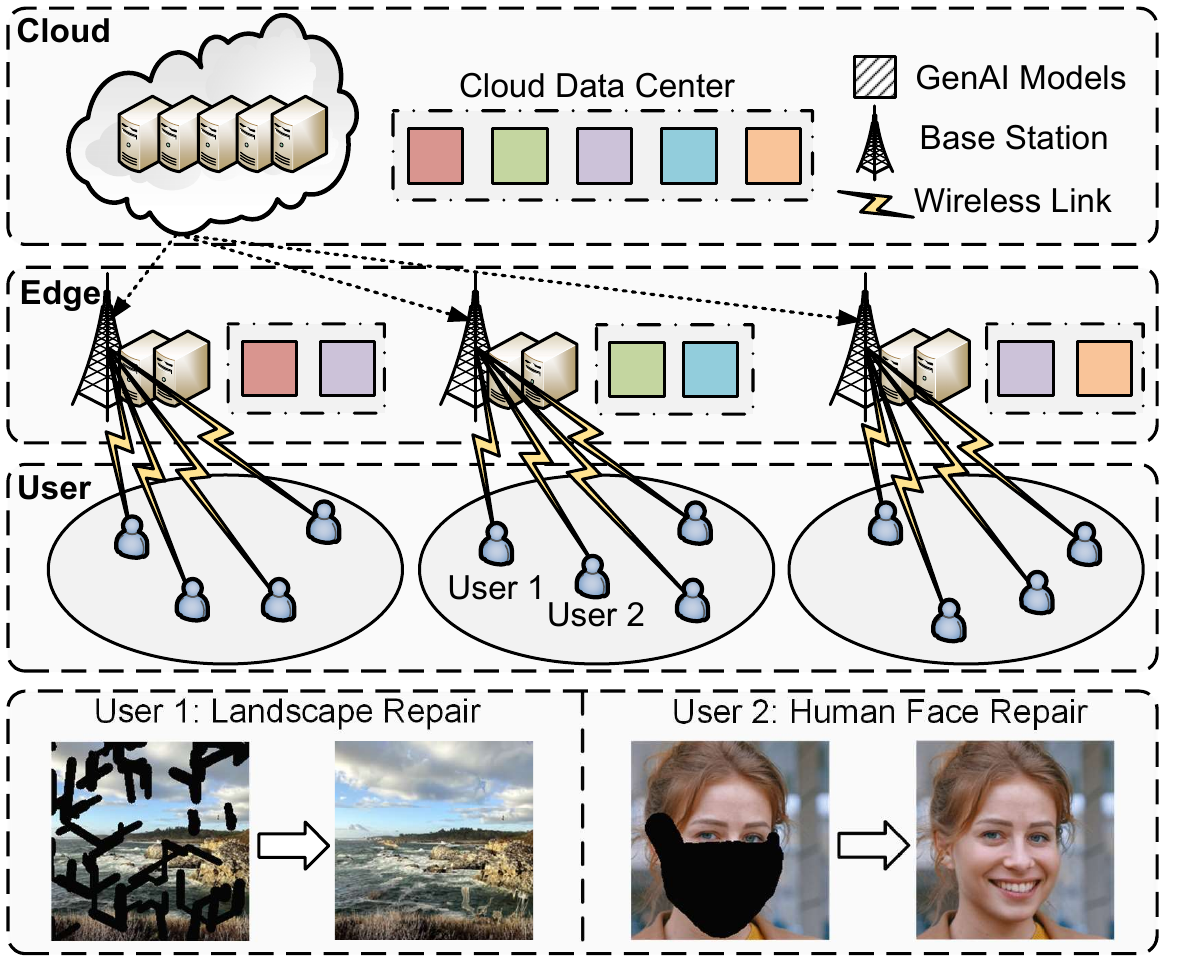}
\centering
\vspace{-1.5mm}
\caption{A schematic of the user-edge-cloud orchestrated architecture for provisioning AIGC services.}
\label{fig:Networks_Architecture}
\vspace{-1mm}
\end{figure}

We consider a three-tier GenAI-enabled mobile network, as shown in Fig.~\ref{fig:Networks_Architecture}, consisting of a cloud data center, a base station (BS) co-located with an edge server,\footnote{Hereafter, we use BS and edge server interchangeably.} and $U$ users denoted by the set $\mathcal{U}=\{1, ...,U\}$. We denote $\mathcal{M}=\{1,...,M\}$ as the set of types of GenAI models, each trained on a unique dataset. The cloud data center, with its high computing power and abundant storage capacity, can be used to train the GenAI models. The edge server, with its limited resources, is responsible for caching a subset of GenAI models and delivering high-quality, low-latency AIGC services to the users in its close proximity.\footnote{In this paper, we consider a single edge server in the system. Similar considerations can also be found in existing works~\cite{9385953,lin2021optimizing,fan2022dnn}. Modeling and analyzing cooperations and competitions among different edge servers for AIGC service provisioning can be further extended by incorporating inter-cell interference and user cell switching.} 

Motivated by the fact that changes in the popularity of AIGC occur more slowly (typically over hours) compared to changes in users' wireless channels (usually within minutes), we adopt a discrete time representation of the system, further divided into two-timescales: a long-timescale (referred to as a \textit{time frame}) for GenAI model caching placement, and a short-timescale (referred to as a \textit{time slot}) for resource allocation. The time frame index is denoted by $t\in \mathcal{T}=\{1,...,T\}$, with each time frame containing $K$ time slots. The time slot index is denoted by $k\in \mathcal{K}=\{1,...,K\}$, each with duration of $\tau$ (in seconds). At each time slot $k$, each user $u$ randomly generates an AIGC service request, characterized by a two-tuple $[\varphi_{u,t}(k),d^{\mathsf{in}}_{u,t}(k)]$, where $\varphi_{u,t}(k)$ denotes the type of AIGC service corresponding to a specific GenAI model in $\mathcal{M}$, and $d^{\mathsf{in}}_{u,t}(k)$ denotes the input data size (in bits). It is assumed that users do not have sufficient storage to cache GenAI models~\cite{10418532} and must either migrate the AIGC service request to the BS (if the corresponding GenAI model is available) or offload to the cloud data center, which stores all GenAI models. For ease of reference, key notations used in the article are summarized in Table~\ref{table3}.

\vspace{-1mm}

\vspace{-2mm}
\subsection{Model Caching}\label{subsec:caching_model}
\vspace{-.15mm}
To formalize the GenAI models considered in this work, we begin by introducing their characteristics. In particular,
GenAI models are DNNs with well-trained parameters that can automate the creation of various types of content through the inference process. We focus on diffusion-based GenAI models, such as \emph{RePaint}, which are trained on different datasets to represent distinct GenAI models. For instance, \emph{RePaint} trained on a dataset of celebrities’ faces can be used to repair corrupted human images, while \emph{RePaint} trained on a landscape scene dataset can generate scenery pictures.\footnote{Accordingly, this paper uses image-generating AIGC services based on diffusion GenAI model as a typical example. However, similar methodologies can be extended to other types of content, such as text and audio.} As a result, an edge server can execute an AIGC service request only when one of its cached GenAI models is trained on the relevant dataset.

At the beginning of each time frame $t$, the BS updates its model caching decisions $\bm{\varrho}(t)=\{\varrho_1(t),\dots,\varrho_m(t),\dots,\varrho_M(t)\}$ and maintains this cache configuration throughout $K$ time slots. Specifically, $\varrho_m(t)=1$ indicates that the $m$-th GenAI model is cached at the BS during the $t$-th time frame; otherwise, $\varrho_m(t)=0$. 
Additionally, we assume that the probability of user $u\in \mathcal{U}$ requesting AIGC service $m\in \mathcal{M}$ at time slot $k$ of time frame $t$ follows a Zipf distribution~\cite{yao2023cooperative}, expressed as
\vspace{-1.5mm}
\begin{equation}\label{eq:Zipf_distribution}
P\{\varphi_{u,t}(k)=m\} = \frac{m^{-\gamma(t)}}{\sum_{i\in\mathcal{M}}i^{-\gamma(t)}},
\vspace{-.5mm}
\end{equation}
where $\gamma(t)$ denotes the skewness of popularity at time frame $t$. Specifically, considering that users' AIGC requests can fluctuate due to dynamic fashion trends, we model the time-varying popularity $\gamma(t)$ as a finite-state Markov sequence~\cite{zhang2022two} $\gamma(t) \in \Gamma=\{\gamma_1,...,\gamma_{J}\}$ with a total of $J$ states/configurations. The transition probability of the skewness of popularity across consecutive time frames is represented by $P\{\gamma(t+1)|\gamma(t)\}$. 

\begin{table}[!t]
\vspace{-2mm}
\centering
\footnotesize
\caption{Summary of Key Notations.}
\label{table3}
\rowcolors{1}{lightblue!30}{white}
\begin{tabular}{|c|p{6.8cm}|}
\hline
\textbf{Notations}                    & \textbf{Description}   \\ \hline \hline
$\mathcal{U}$                  & Index set of users                \\ \hline
$\mathcal{M}$                    & Index set of GenAI models               \\ \hline
$\mathcal{T}$          & Index set of time frames       \\ \hline
$\mathcal{K}$   & Index set of time slots          \\ \hline
$\varphi_{u,t}(k)$                 & AIGC service type of user $u$ generated at time slot $k$ of time frame $t$   \\ \hline
$d^{\mathsf{in}}_{u,t}(k)$                  & Input data size of $\varphi_{u,t}(k)$-th type AIGC service                \\ \hline
$R^{\mathsf{up}}_{u,t}(k)$                  & Uplink transmission rate from user $u$ to the BS at time slot $k$ of time frame $t$      \\ \hline
$D^{\mathsf{up}}_{u,t}(k)$                  & Uplink transmission delay for user $u$ to
migrate its AIGC service request at time slot $k$ of time frame $t$       \\ \hline
$R^{\mathsf{dw}}_{u,t}(k)$                  & Downlink transmission rate from the BS to user $u$ at time slot $k$ of time frame $t$          \\ \hline
$D^{\mathsf{dw}}_{u,t}(k)$              &  Feedback delay for user $u$ at time slot $k$ of time frame $t$           \\ \hline
$\mathcal{L}$         &  Total number of denoising steps performed at the edge server                \\ \hline
$B^{\mathsf{gt}}_{u,t}(k)$    &  TV value of the generated image perceived by user $u$ at time slot
$k$ of time frame $t$            \\ \hline
$D^{\mathsf{gt}}_{u,t}(k)$  & Image generation time for user $u$ at time slot $k$ of time frame $t$              \\ \hline
$D^{\mathsf{tl}}_{u,t}(k)$                  & AIGC service provisioning delay for user $u$ at time slot $k$ of time frame $t$          \\ \hline
$\varrho_m(t)$                  & GenAI model caching decision at time frame $t$         \\ \hline
$b_{u,t}(k)$                          & Bandwidth allocation ratio for user $u$ at time slot $k$ of time frame $t$       \\\hline 
$\xi_{u,t}(k)$                          & Denoising step allocation ratio for user $u$ at time slot $k$ of time frame $t$        \\\hline
\end{tabular}
\end{table}

\vspace{-2mm}
\subsection{Communication Model}\label{subsec:comm_model}
\vspace{-.15mm}
Since users must both send their AIGC requests to the BS and retrieve the corresponding images, we formulate the service migration and result feedback models below.

\subsubsection{Service Migration} 
We model the transmission rate from user $u$ to the BS at time slot $k$ of time frame $t$ as follows:
\vspace{-1mm}
\begin{equation}\label{eq:uplink_transrate}
R^{\mathsf{up}}_{u,t}(k)= b_{u,t}(k)W^{\mathsf{up}}\text{log}_2\Big(1+\frac{p_uh_{u,t}(k)}{N_0b_{u,t}(k)W^{\mathsf{up}}} \Big),
\vspace{-.5mm}
\end{equation}
where $b_{u,t}(k)$ denotes the \textit{bandwidth allocation ratio} for user $u$, $W^{\mathsf{up}}$ (in Hz) represents the uplink channel bandwidth, $p_u$ is the transmit power of user $u$, and $N_0$ is the noise power spectral density. Additionally, $h_{u,t}(k)= g_{u,t}(k)|\delta_{u,t}(k)|^2$~\cite{9562522} represents the channel gain, capturing both path loss and signal fading. Specifically, $\delta_{u,t}(k) \sim \mathcal{CN}(0,1)$ is the Rayleigh fading component, varying i.i.d. across different time slots, and the path loss $g_{u,t}(k)$ (in dB) is modeled as
\vspace{-1.5mm}
\begin{equation}\label{eq:path_loss}
g_{u,t}(k)=-128.1-37.6\text{log}_{10}\textrm{dis}_{u,t}(k),
\vspace{-.5mm}
\end{equation}
where $\textrm{dis}_{u,t}(k)$ denotes the Euclidean distance between user $u$ and the BS at time slot $k$ of time frame $t$. Users' location distribution at each time slot $k$ of time frame $t$, denoted by $\lambda_t(k)$, is assumed to come from a finite set $\lambda_t(k) \in \Lambda=\{\lambda_1, ..., \lambda_I\}$, with a total of $I$ states~\cite{zhang2022two}. The transition probability of the users' location distribution between consecutive time slots in each time frame $t$ is represented by $P\{\lambda_t(k+1)|\lambda_t(k)\}$. 

Consequently, the uplink transmission delay for user $u$ to migrate its AIGC service request at time slot $k$ of time frame $t$, denoted by $D^{\mathsf{up}}_{u,t}(k)$, can be calculated as follows:
\vspace{-1mm}
\begin{equation}\label{eq:uplink_transmission_delay}
D^{\mathsf{up}}_{u,t}(k)\!=\! \left\{ \begin{array}{l}
            {d^{\mathsf{in}}_{u,t}(k)}/{R^{\mathsf{up}}_{u,t}(k)}, \quad \text{if}\ \varrho_{m}(t)=1, m=\varphi_{u,t}(k) \\
			{d^{\mathsf{in}}_{u,t}(k)}/{R^{\mathsf{up}}_{u,t}(k)}+{d^{\mathsf{in}}_{u,t}(k)}/{R^{\mathsf{bc}}}, \quad \text{otherwise},
		\end{array} \right.
\vspace{-.5mm}
\end{equation}
where $\varrho_{m}(t)=1, m=\varphi_{u,t}(k)$ indicates that the GenAI model corresponding to the AIGC service requested by user $u$ at time slot $k$ is cached at the BS, allowing the AIGC service to be provided at the edge. Otherwise, the AIGC service will be migrated to the cloud data center, with $R^{\mathsf{bc}}$ denoting the fixed wired transmission rate between the BS and the cloud, which determines the additional backhaul transmission delay.

\subsubsection{Result Feedback} After the AIGC is generated, the result feedback delay can be calculated as the AIGC retrieval delay for user $u$. Similar to \eqref{eq:uplink_transmission_delay}, the transmission rate from the BS to user $u$ at time slot $k$ of time frame $t$ is given by~\cite{park2020mobile}
\vspace{-1mm}
\begin{equation}\label{eq:uplink_transrate}
R^{\mathsf{dw}}_{u,t}(k)= W^{\mathsf{dw}}\text{log}_2\Big(1+\frac{p^{\mathsf{BS}}h_{u,t}(k)}{N_0W^{\mathsf{dw}}} \Big),
\vspace{-.75mm}
\end{equation}
where $p^{\mathsf{BS}}$ is the transmit power of the BS, and $W^{\mathsf{dw}}$ (in Hz) represents the per-user downlink channel bandwidth. Consequently, the result feedback delay for user $u$ when downloading the generated image at time slot $k$ of time frame $t$ can be calculated as follows:
\vspace{-1mm}
\begin{equation}\label{eq:downlink_transmission_delay}
\hspace{-3mm}
D^{\mathsf{dw}}_{u,t}(k)= \left\{ \begin{array}{l}
            {d^{\mathsf{op}}_{m}}/{R^{\mathsf{dw}}_{u,t}(k)}, \quad \text{if}\ \varrho_{m}(t)=1, m=\varphi_{u,t}(k) \\[.3em]
			{d^{\mathsf{op}}_{m}}/{R^{\mathsf{dw}}_{u,t}(k)}+{d^{\mathsf{op}}_{m}}/{R^{\mathsf{cb}}}, \quad \text{otherwise},
		\end{array} \right.
\vspace{-.5mm}
\hspace{-3mm}
\end{equation}
where $d^{\mathsf{op}}_{m}$ is the output data size (in bits) of $m$-th type AIGC service (e.g., the size of the reconstructed image) dictated by the architecture of its GenAI model, and $R^{\mathsf{cb}}$ is the fixed wired transmission rate between the cloud data center and the BS.


\vspace{-1mm}
\begin{figure*}[t!]
\vspace{-2mm}
\includegraphics[width=.98\textwidth]{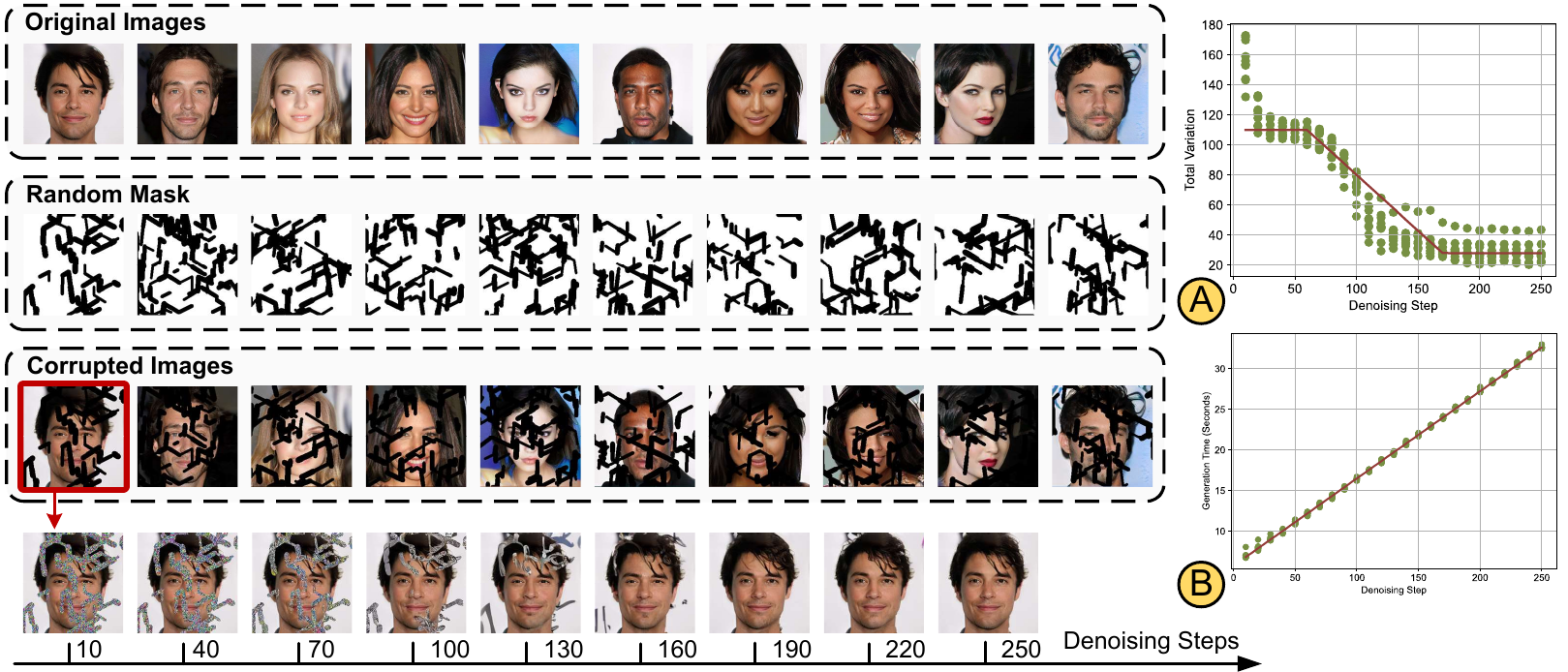}
\centering
\vspace{-1.5mm}
\caption{An example of an edge-enabled AIGC service for restoring corrupted images of human faces.}
\label{fig:Fitting_Relationship}
\end{figure*}

\begin{table*}[!t]

  \hrule
   \vspace{0.5mm}
   \begin{equation}\label{eq:image_generation_quality}
 B^{\mathsf{gt}}_{u,t}(k)= \left\{ \begin{array}{l}
            A_2,  \quad\quad \xi_{u,t}(k)\mathcal{L}\leq A_1  \\
            \frac{A_4-A_2}{A_3-A_1}(\xi_{u,t}(k)\mathcal{L}-A_1)+A_2, \quad A_1< \xi_{u,t}(k)\mathcal{L}< A_3 \\
			A_4, \quad\quad \xi_{u,t}(k)\mathcal{L} \geq A_3
		\end{array} \right., \ \varrho_{m}(t)=1, m=\varphi_{u,t}(k)
 \end{equation}
    \vspace{-0.5mm}
 \hrule
 \vspace{-3mm}
\end{table*}

\vspace{-3mm}
\subsection{Computing Model}\label{subsec:comp_model}
\vspace{-.15mm}
Before introducing our computing model, we first highlight the major challenges in evaluating the performance of AIGC services. Specifically, unlike existing works~\cite{9861697,9090328,10076905,10609797}, which focus on improving performance indicators such as task execution delay and energy consumption using explicit mathematical models, other evaluation metrics for AIGC services are challenging to quantify without a mathematical expression. To tackle this, we focus on a specific AIGC, \textit{images}, where achieving optimal resource allocation schemes for high-quality, low-latency AIGC services requires modeling the mathematical functions between computational resources (e.g., denoising steps in diffusion models) and both image generation delay and the quality of the generated image. 
To this end, we conduct experiments with practical GenAI models and use a fitting method to develop one of the first mathematical functions between GenAI image generation quality, image generation delay, and computational resource consumption.

\subsubsection{Image Generation Quality} Given that image quality is inherently subjective, we focus on an image-based metric to capture the image quality by modeling key physiological and psychovisual features of the human visual system. In particular, we leverage total variation (TV)~\cite{kastryulin1pytorch}, a metric recognized for its effectiveness in measuring image smoothness, to assess the perceived quality of AI-generated images. Specifically, TV quantifies `roughness' or `discontinuity' by summing the absolute differences between adjacent pixels in an image. A lower TV value indicates higher image quality.\footnote{Other image-based metrics, such as BRISQUE\cite{mittal2012no}, DSS~\cite{gatys2015neural}, and MDSI~\cite{kastryulin1pytorch}, are also applicable for evaluating perceived image quality, and we use TV as a representative example in this paper.}

We conduct experiments to explore the mathematical relationship between the number of denoising steps, which capture the inference process of the diffusion-based GenAI model, and image quality as measured by TV. As illustrated in Fig.~\ref{fig:Fitting_Relationship}, we first generated 10 corrupted images of human faces using random masks, which were then inpainted. These corrupted images were processed using \emph{RePaint}, and we observed that they are gradually recovered as the number of denoising steps increases. Additionally, we present the TV values of these 10 corrupted images at different denoising steps (part A in Fig.~\ref{fig:Fitting_Relationship}) and fit a mathematical function using a piecewise function (highlighted in red). 


As a result, we propose a general model relating the TV value of the generated image perceived by user $u$ at time slot $k$ of time frame $t$, denoted by $B^{\mathsf{gt}}_{u,t}(k)$, to the allocated computational resource (i.e., denoising steps), given by~\eqref{eq:image_generation_quality} at the top of this page. Specifically, $\xi_{u,t}(k) \in [0,1]$ represents the denoising step allocation ratio for user $n$ at time slot $k$, and $\mathcal{L}$ denotes the total number of denoising steps performed at the edge server.\footnote{Although denoising steps in conventional diffusion models are represented as integers, in this paper, we relax them into proportional allocations at the level of computational resource scheduling. This approach makes the allocation of computational resources more flexible and manageable.} Also,~\eqref{eq:image_generation_quality} includes four parameters: $A_1=60$, representing the minimum number of denoising steps where image quality begins to improve; $A_2=110$, indicating the lower bound of image quality; $A_3=170$, marking the number of denoising steps when image quality starts to stabilize; $A_4=28$, denoting the highest image quality value. When the GenAI model corresponding to the AIGC service requested by user $u$ at time slot $k$ is not cached at the edge server, i.e., $\varrho_{m}(t)\neq1, m=\varphi_{u,t}(k)$, we consider $ B^{\mathsf{gt}}_{u,t}(k)=A_4$, as the cloud has sufficient computing resources to generate an image with the highest quality. This modeling technique and methodology has the potential to be applied to a wider range of future problems on AIGC delivery in wireless networks.

\subsubsection{Image Generation Delay} Following the same methodology, we present the generation times of these 10 recovered images at different denoising steps (part B in Fig.~\ref{fig:Fitting_Relationship}) and fit a function, which appears to be a linear function (highlighted in red). As a result, we propose a model relating the image generation time for user $u$ at time slot $k$ of time frame $t$, denoted by $D^{\mathsf{gt}}_{u,t}(k)$, to the allocated denoising steps as follows:
\vspace{-1.5mm}
\begin{align}\label{eq:image_generation_delay}
    D^{\mathsf{gt}}_{u,t}(k) = B_1\xi_{u,t}(k)\mathcal{L}+B_2, \ \varrho_{m}(t)=1, m=\varphi_{u,t}(k),
    \vspace{-.5mm}
\vspace{-.5mm}
\end{align}
which includes two parameters: $B_1=0.18$ and $B_2=5.74$. When the GenAI model corresponding to the AIGC service requested by user $u$ at time slot $k$ is not cached at the edge server, i.e., $\varrho_{m}(t)\neq1, m=\varphi_{u,t}(k)$, we consider $ D^{\mathsf{gt}}_{u,t}(k)=B_1A_3+B_2$, where the computing resources allocated by the cloud are defined as the minimum threshold required to generate the image at the highest quality (i.e., $A_3$). In summary, considering~\eqref{eq:uplink_transmission_delay},~\eqref{eq:downlink_transmission_delay}, and~\eqref{eq:image_generation_delay}, the AIGC service provisioning delay for user $u$ at time slot $k$ of time frame $t$ is 
\vspace{-0.5mm}
\begin{align}\label{eq:total_delay}
    D^{\mathsf{tl}}_{u,t}(k) = D^{\mathsf{up}}_{u,t}(k)+D^{\mathsf{dw}}_{u,t}(k)+D^{\mathsf{gt}}_{u,t}(k).
\vspace{-5mm}
\end{align}

Consequently, we form a utility function via the weighted sum of generation delay and the TV value of the generated image for user $u$ at time slot $k$ of time frame $t$ as 
\vspace{-0.75mm}
\begin{align}\label{eq:weighted_sum}
    G_{u,t}(k) = \alpha D^{\mathsf{tl}}_{u,t}(k)+(1-\alpha)B^{\mathsf{gt}}_{u,t}(k),
    \vspace{-.5mm}
\vspace{-2mm}
\end{align}
capturing the trade-off between the quality and latency of AIGC services. Here, $\alpha \in [0,1]$ is a preference weight factor.

\vspace{-1mm}
\section{Problem Formulation} \label{sec:probblem_formulation}
\vspace{-.7mm}
We now formulate the problem of two-timescale model caching and resource allocation as a dynamic long-term optimization. Our goal is to minimize the utility in~\eqref{eq:weighted_sum} for all users across all time frames and time slots. This involves both long-timescale GenAI modeling caching scheduling and short-timescale allocation of computing and communication resources. Mathematically, we define this problem as $\mathbf{P1}$ below:
\vspace{-1.5mm}
\begin{align}
     \hspace{-3mm}&\mathbf{P1}: \quad\quad\quad \min_{\bm{\varrho},\bm{b}, \bm{\xi}} \frac{1}{TKU}\sum_{t\in\mathcal{T}}\sum_{k\in\mathcal{K}}\sum_{u\in\mathcal{U}} G_{u,t}(k)\hspace{-40mm}\label{eq:problem1} \\[-.45em]
     &\hspace{-3mm} \textrm{s.t.}\nonumber\\ 
     &\hspace{-3mm}  \varrho_m(t) \in \{0,1\}, \ \forall t\in \mathcal{T}, m\in \mathcal{M}, \tag{11a} \label{eq:Caching_Constraint}\\[-.2em]
     &\hspace{-3mm}  b_{u,t}(k) \in [0,1], \ \forall t\in \mathcal{T}, k\in \mathcal{K}, u\in\mathcal{U}, \tag{11b} \label{eq:Comm_Constraint}\\[-.2em]
     &\hspace{-3mm}  \xi_{u,t}(k) \in [0,1], \ \forall t\in \mathcal{T}, k\in \mathcal{K}, u\in\mathcal{U}, \tag{11c} \label{eq:Comp_Constraint}\\[-.2em] 
     &\hspace{-3mm}  \sum_{m \in \mathcal{M}}\varrho_m(t)c_m \leq C,  \ \forall t\in \mathcal{T}, \tag{11d} \label{eq:Caching_Resource_Constraint}\\[-.2em]
     &\hspace{-3mm}   \sum_{u \in \mathcal{U}}b_{u,t}(k)\leq 1,  \ \forall t\in \mathcal{T}, k\in \mathcal{K}, \tag{11e}\label{eq:Comm_Resource_Constraint}\\[-.2em]
     &\hspace{-3mm}  \sum_{u \in \mathcal{U}}\xi_{u,t}(k)\leq 1,  \ \forall t\in \mathcal{T}, k\in \mathcal{K}, \tag{11f} \label{eq:Comp_Resource_Constraint}\\[-.2em]
     &\hspace{-3mm}  \xi_{u,t}(k) \leq \varrho_{m}(t), m=\varphi_{u,t}(k),  \ \forall t\in \mathcal{T}, k\in \mathcal{K}, u\in \mathcal{U}, \tag{11g} \label{eq:Variable_Constraint}\\[-.2em]
     &\hspace{-3mm}  D^{\mathsf{tl}}_{u,t}(k) \leq \tau,  \ \forall t\in \mathcal{T}, k\in \mathcal{K}, u\in \mathcal{U}, \tag{11h} \label{eq:Duration_Constraint}
     \vspace{-1.5mm}
\end{align}
where $\bm{\varrho}=\{\varrho_m(t)\}_{m\in\mathcal{M},t\in \mathcal{T}}$ is the caching decision vector for GenAI models, impacting the latency of GenAI content requests and receptions as  in~\eqref{eq:uplink_transmission_delay} and~\eqref{eq:downlink_transmission_delay}, $\bm{b}=\{b_{u,t}(k)\}_{u \in \mathcal{U},t\in \mathcal{T}, k \in \mathcal{K}}$ is the communication resource allocation ratio vector for users, determining the bandwidth allocation as  in~\eqref{eq:uplink_transrate}, and $\bm{\xi}=\{\xi_{u,t}(k)\}_{u \in \mathcal{U},t\in \mathcal{T}, k \in \mathcal{K}}$ is the computing resource allocation ratio vector for users, which influence both the quality and latency of AIGC generation, as in~\eqref{eq:image_generation_quality} and~\eqref{eq:image_generation_delay}, respectively.

In $\mathbf{P1}$, constraint~\eqref{eq:Caching_Constraint} ensures that the model caching decision is binary. Constraints~\eqref{eq:Comm_Constraint} and \eqref{eq:Comp_Constraint} define the value ranges for the communication and computation resources allocated by the BS to different users. Constraints~\eqref{eq:Caching_Resource_Constraint}-\eqref{eq:Comp_Resource_Constraint} specify the limitations on the BS's caching, bandwidth, and computing capacities, respectively. In constraint~\eqref{eq:Caching_Resource_Constraint}, $c_m$ represents the storage requirement (in GB) for the $m$-th type GenAI model, while $C$ denotes the maximum storage capacity (in GB) of the BS. Constraint~\eqref{eq:Variable_Constraint} indicates that the BS's computing resources will not be allocated for the $m$-th type AIGC service if the relevant GenAI model is not cached at the BS. Additionally, constraint~\eqref{eq:Duration_Constraint} ensures that the AIGC service provisioning delay of user $u$ at time slot $k$ of time frame $t$ given by~\eqref{eq:total_delay} does not exceed the duration of the time slot, thereby preventing potential computation backlogs when AIGC service requests periodically arrive from the users.

\vspace{-1mm}
\begin{remark}\label{rem:NPhard} The objective function in~\eqref{eq:problem1} captures a long-term optimization problem, involving a sum over multiple timescales and users, highlighting its dynamic nature. Additionally, the calculation of image generation quality, formulated as a piecewise function in~\eqref{eq:image_generation_quality} and included in the objective function~\eqref{eq:problem1}, further introduces nonlinearity. Furthermore, due to the presence of both continuous and binary variables, as dictated by~\eqref{eq:Caching_Constraint}-\eqref{eq:Comp_Constraint}, problem $\mathbf{P1}$ is a mixed integer nonlinear program (MINLP), which is generally NP-hard. As a result, efficiently solving problem $\mathbf{P1}$ is challenging.
\end{remark}
\vspace{-.5mm}

\vspace{-4mm}
\subsection{Problem Decomposition across Different Timescales}
Since GenAI model updates typically occur only after the collection of substantial new data~\cite{zhang2022two} while AIGC service requests arrive within seconds from users, in the following, we decompose the original problem $\mathbf{P1}$ to facilitate its solution.

\subsubsection{Resource Allocation Subproblem in Short-Timescale} From the short-timescale perspective (at each time slot $k$), our goal is to minimize $G_{u,t}(k)$ for all users across all time slots, given the GenAI model caching decisions made for the relevant time frame $t$, by determining the computing and communication resource allocation. Subsequently, we mathematically formulate the resource allocation subproblem $\mathbf{P2}$  as follows:
\vspace{-0.25mm}
\begin{align}
     \hspace{-3mm}\mathbf{P2}:\quad\quad\quad&  \min_{\bm{b}, \bm{\xi}}\frac{1}{KU} \sum_{k\in\mathcal{K}}\sum_{u\in\mathcal{U}} G_{u,t}(k)\label{eq:problem2} \\
     & \textrm{s.t.} \quad \eqref{eq:Comm_Constraint}, \eqref{eq:Comp_Constraint}, \eqref{eq:Comm_Resource_Constraint}-\eqref{eq:Duration_Constraint}.  \nonumber
     \vspace{-1.75mm}
\end{align}

\subsubsection{Model Caching Subproblem in Long-Timescale} From the long-timescale perspective (at each time frame $t$), our goal is to obtain the GenAI model caching decisions that minimize $G_{u,t}(k)$ for all users across all time frames. We thus  formulate the model caching subproblem $\mathbf{P3}$ as follows:
\vspace{-0.5mm}
\begin{align}
     \hspace{-3mm}\mathbf{P3}:\quad&  \min_{\bm{\varrho}}\frac{1}{TKU}\sum_{t\in\mathcal{T}}\sum_{k\in\mathcal{K}}\sum_{u\in\mathcal{U}} G_{u,t}(k)\label{eq:problem3} \\
     & \textrm{s.t.} \quad \eqref{eq:Caching_Constraint}, \eqref{eq:Caching_Resource_Constraint}.  \nonumber
     \vspace{-1.5mm}
\end{align}
In the following, we introduce a two-timescale deep reinforcement learning (T2DRL) algorithm to  solve $\mathbf{P2}$ and $\mathbf{P3}$.

\vspace{-3mm}
\section{Basic Idea of Diffusion Models} \label{sec:diffusion_model}
Before delving into T2DRL, we first explain our motivation for integrating the diffusion model with DRL (i.e., diffusion-based deep deterministic policy gradient algorithm). We then detail how the diffusion model is customized to generate decisions for communication and computing resource allocation.

\vspace{-6mm}
\subsection{Motivation of Adopting Diffusion Model}  \label{subsec:DDPM_motivation} 
In addition to the shortcomings of the multi-layer perceptron (MLP) commonly used in conventional DRL algorithms, as detailed in Sec.~\ref{subsec:DRL_optimization}, our motivation for utilizing diffusion models arises from their unique capability to integrate with the DRL framework. 
Specifically, in a conventional diffusion model, a user can input a text prompt (e.g., `an apple on the table') to guide the model in generating a corresponding image. In our scenario, we consider optimal communication and computing resource allocation decisions as the `image' we aim to generate through the diffusion model, with dynamic environmental information—such as time-varying channel conditions, user movement, and AIGC service popularity—acting as the `text prompt' guiding the training process. After training, diffusion models can generate optimal resource allocation decisions tailored to any dynamic wireless environment condition~\cite{10529221}. This adaptive ability to generate solutions is particularly valuable for the dynamic environment considered in this paper.


\vspace{-3mm}
\subsection{Preliminaries of Diffusion Models}  \label{subsec:DDPM} 
Based on the denoising diffusion probabilistic model (DDPM)~\cite{ho2020denoising}, conventional applications of diffusion models were originally designed for image generation. Specifically, DDPM involves two critical processes during training: the \emph{forward process}, where noise is progressively added to the original image at each noising step until it becomes indistinguishable from pure Gaussian noise; the \emph{reverse process}, where the noise learned by an MLP is gradually removed at each denoising step to recover the original image. 

We first encapsulate the optimal communication resource allocation $\bm{b}^*_t(k)= \{b_{1,t}(k),...,b_{U,t}(k)\}$ and the optimal computing resource allocation $\bm{\xi}^*_{t}(k) = \{\xi_{1,t}(k),...,\xi_{U,t}(k)\}$ at time slot $k$  into a vector $\bm{x}^0_t(k) = \{\bm{b}^*_t(k), \bm{\xi}^*_{t}(k)\}$, which is considered the ``original policy'' for DDPM. The corresponding forward and reverse processes are then detailed below.

\begin{figure}[t]
\vspace{-1mm}
\includegraphics[width=.48\textwidth]{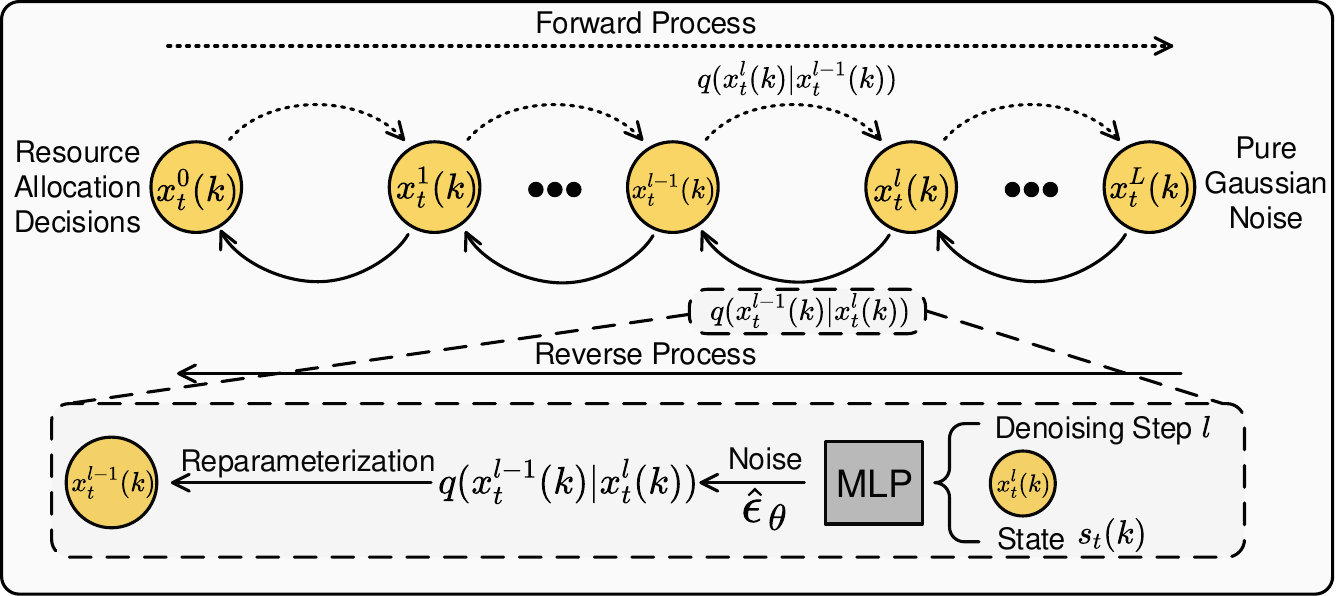}
\centering
\caption{An illustration of the diffusion model tailored to generate optimal decisions for communication and computing resource allocation at time slot $k$.}
\label{fig:DDPM}
\end{figure}
\subsubsection{Forward Process}  \label{subsubsec:forward_process}  
A schematic of the diffusion model tailored to generating optimal communication and computing resource allocation decisions at time slot $k$ is depicted in Fig.~\ref{fig:DDPM}. Specifically, the forward process is modeled as a Markov chain with $L$ noising/denoising steps. Starting from the optimal solution $\bm{x}^0_t(k)$, at each noising step $l$, the forward process adds Gaussian noise to $\bm{x}^{l-1}_t(k)$ to yield $\bm{x}^{l}_t(k)$. The transition is defined as a normal distribution with a mean of $\sqrt{1-\beta_l}\bm{x}^{l-1}_t(k)$ and a variance of $\beta_l\mathbf{I}$ given by
\vspace{-1.05mm}
\begin{align}\label{eq:forward_distribution}
    q(\bm{x}^l_t(k)|\bm{x}^{l-1}_t(k))= \mathcal{N}(\bm{x}^l_t(k);\sqrt{1-\beta_l}\bm{x}^{l-1}_t(k),\beta_l\mathbf{I}),
    \vspace{-1.5mm}
\end{align}
where $\beta_l$ is the diffusion rate at noising step $l$~\cite{ho2020denoising}, calculated as $\beta_l=1- e^{-\frac{\beta^{\mathsf{min}}}{L}-\frac{2l-1}{2L^2}(\beta^{\mathsf{max}}-\beta^{\mathsf{min}})}$, with $\beta^{\mathsf{min}}$ and $\beta^{\mathsf{max}}$ being the predetermined minimum and maximum diffusion rates, respectively, and $\mathbf{I}$ representing the identity matrix.

\begin{remark}
   Note that in this paper, the ``denoising step" serves dual purposes: (P-i) as computational resources allocated to different users for AIGC (i.e., image) generation, and (P-ii) as a parameter in D3PG that guides the training process. In (P-i), the diffusion model is treated as a typical GenAI model for generating AIGC, while in (P-ii) it is regarded as a new component to enhance the capabilities of conventional DRL.
\end{remark}

Based on~\eqref{eq:forward_distribution}, since $\bm{x}^l_t(k) \sim \mathcal{N}(\sqrt{1-\beta_l}\bm{x}^{l-1}_t(k),\beta_l\mathbf{I})$, the mathematical relationship between $\bm{x}^{l-1}_t(k)$ and $\bm{x}^l_t(k)$ can be derived using the reparameterization technique as follows~\cite{ho2020denoising}:
\vspace{-1.5mm}
\begin{equation}\label{eq:forward_update}
    \bm{x}^l_t(k)= \sqrt{1-\beta_l}\bm{x}^{l-1}_t(k)+\sqrt{\beta_l}\bm{\epsilon}_{l-1},
    \vspace{-1.5mm}
\end{equation}
where $\bm{\epsilon}_{l-1}$ is Gaussian noise sampled from $\mathcal{N}(0,\mathbf{I})$. Finally, based on~\eqref{eq:forward_update}, the relationship between $\bm{x}^0_t(k)$ and $\bm{x}^l_t(k)$ at any noising step $l$ can be calculated as follows\footnote{Since $\mathbf{P1}$ is an MINLP, it is challenging to obtain the optimal solution $\bm{x}^0_t(k)$, and thus the forward process is not performed in this work and depicted via dotted lines in Fig.~\ref{fig:DDPM}. It serves solely to establish the relationship between $\bm{x}^0_t(k)$ and $\bm{x}^l_t(k)$, which is essential for the reverse process described later.}:
\vspace{-1.05mm}
\begin{align}\label{eq:forward_relationship}
    \bm{x}^l_t(k)=\sqrt{\bar{\alpha}_l}\bm{x}^{0}_t(k)+\sqrt{1-\bar{\alpha}_l}\tilde{\bm{\epsilon}}_l,
    \vspace{-1.5mm}
\end{align}
where $\bar{\alpha}_l=\prod_{\ell=1}^{l}\alpha_\ell$ is the cumulative product of $\alpha_\ell$ over the previous noising steps $l$, $\alpha_\ell=1-\beta_\ell$, and $\tilde{\bm{\epsilon}}_l \sim \mathcal{N}(0,\mathbf{I})$. 

\subsubsection{Reverse Process}  \label{subsubsec:reverse_process} 
Based on~\eqref{eq:forward_relationship}, we observe that when $L$ is large, $\bm{x}^L_t(k)$ approximates an isotropic Gaussian distribution. Consequently, in the reverse process, starting from a sample $\bm{x}^L_t(k) \sim \mathcal{N}(0, \mathbf{I})$, a denoiser $\pi_{\bm{\theta}}$, parameterized by $\bm{\theta}$, at each denoising step $l$, takes $\bm{x}^{l}_t(k)$, the current denoising step $l$, and the state $\bm{s}_t(k)$ (to be formulated later in Sec. \ref{subsubsec:D3PG_MDP}), and predicts the noise to be removed at the current denoising step to yield $\bm{x}^{l-1}_t(k)$. The transition from $\bm{x}^{l}_t(k)$ to $\bm{x}^{l-1}_t(k)$ has been shown to follow a Gaussian distribution as follows~\cite{ho2020denoising}:
\vspace{-1.05mm}
\begin{align}\label{eq:reverse_distribution}
q(\bm{x}^{l-1}_t(k)|\bm{x}^l_t(k))= \mathcal{N}(\bm{x}^{l-1}_t(k);\bm{\mu}^l_t(k),\bar{\beta}_l\mathbf{I}),
\vspace{-1.5mm}
\end{align}
where $\bar{\beta}_l=\frac{1-\bar{\alpha}_{l-1}}{1-\bar{\alpha}_l}\beta_l$. Also, the mean $\bm{\mu}^l_t(k)$ can be obtained using the Bayesian formula as follows:
\vspace{-1.05mm}
\begin{align}\label{eq:reverse_original_mean}
\bm{\mu}^l_t(k)= \frac{\sqrt{\alpha_l}(1-\bar{\alpha}_{l-1})}{1-\bar{\alpha}_l}\bm{x}^{l}_t(k)+\frac{\sqrt{\bar{\alpha}_{l-1}}\beta_l}{1-\bar{\alpha}_l}\bm{x}^{0}_t(k).
\vspace{-1.5mm}
\end{align}

Next, by incorporating~\eqref{eq:forward_relationship} into~\eqref{eq:reverse_original_mean}, we can replace $\bm{x}^{0}_t(k)$ with $\bm{x}^{l}_t(k)$, and the mean $\bm{\mu}^l_t(k)$ is reconstructed as follows:
\vspace{-1.05mm}
\begin{align}\label{eq:reverse_mean}
&\bm{\mu}_{\bm{\theta}}(\bm{x}^{l}_t(k),l,\bm{s}_t(k))\nonumber\\
&=\frac{1}{\sqrt{\alpha_l}}\Big[\bm{x}^{l}_t(k)-\frac{1-\alpha_l}{\sqrt{1-\bar{\alpha}_l}}\bm{\hat{\epsilon}}_{\bm{\theta}}(\bm{x}^{l}_t(k),l,\bm{s}_t(k))\Big],
\vspace{-1.5mm}
\end{align}
where $\bm{\hat{\epsilon}}_{\bm{\theta}}(\bm{x}^{l}_t(k),l,\bm{s}_t(k))$ represents the noise predicted by the denoiser $\pi_{\bm{\theta}}$ at denoising step $l$. In DDPM, the mean squared error between the noise $\bm{\epsilon}_l$ introduced during the forward process and the noise $\bm{\hat{\epsilon}}_{\bm{\theta}}$ predicted by the denoiser $\pi_{\bm{\theta}}$ during the reverse process at the corresponding noising step is continually minimized. \emph{However, since the forward process is not performed in this paper, the training objective of the reverse process shifts towards minimizing the objective function given in~\eqref{eq:problem2} in an exploratory manner (as detailed in Sec.~\ref{subsubsec:D3PG_architecture}).}

\vspace{-1mm}
\begin{figure*}[t!]
\includegraphics[width=.98\textwidth]{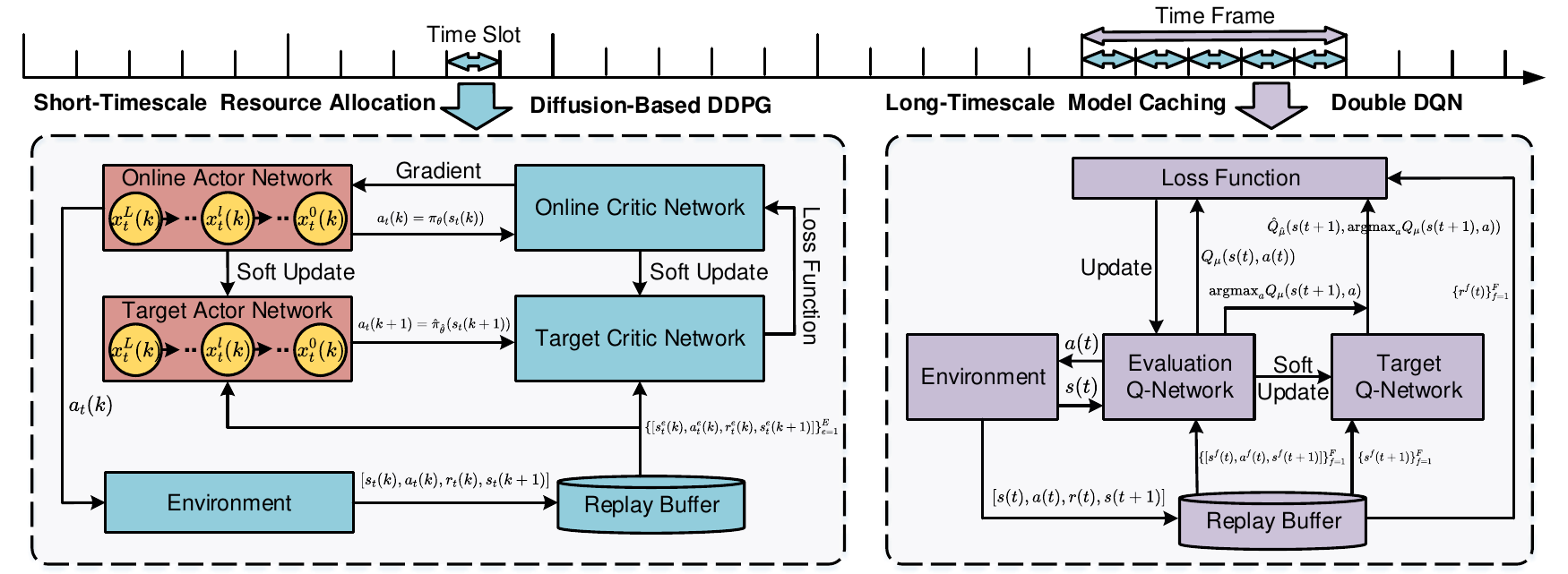}
\centering
\caption{Overall flowchart of the proposed T2DRL algorithm, consisting of the DDQN algorithm operating on the long-timescale and the D3PG algorithm operating on the short-timescale.}
\label{fig:T2DRL_Framework}
\vspace{-4mm}
\end{figure*}
\vspace{3mm}

Finally, based on~\eqref{eq:reverse_distribution}, the mathematical relationship between $\bm{x}^l_t(k)$ and $\bm{x}^{l-1}_t(k)$ can be derived using the reparameterization technique as follows:
\vspace{-1.5mm}
\begin{equation}\label{eq:reverse_update}
    \bm{x}^{l-1}_t(k)= \bm{\mu}_{\bm{\theta}}(\bm{x}^{l}_t(k),l,\bm{s}_t(k))+\sqrt{\bar{\beta}_l}\bar{\bm{\epsilon}}_l,
    \vspace{-1.5mm}
\end{equation}
where $\bar{\bm{\epsilon}}_l \sim \mathcal{N}(0,\mathbf{I})$. In this paper, the denoiser $\pi_{\bm{\theta}}$ is considered the optimal decision generation network. By iteratively applying the rule  in~\eqref{eq:reverse_update}, we obtain the optimal communication and computing resource allocation $\bm{x}^0_t(k) = \{\bm{b}^*_t(k), \xi^*_{t}(k)\}$ after $L$ denoising steps (detailed in Algorithm~\ref{algo：T2DRL}).

\vspace{-3.5mm}
\section{T2DRL for Model Caching and Resource Allocation}\label{sec:T2DRL}
\vspace{-.15mm}
In the following, we first provide an overview of our proposed T2DRL algorithm. Next, we detail its architecture and its components. Finally, we analyze the computational complexity associated with the proposed method.


\vspace{-3mm}
\subsection{Overview of the Proposed T2DRL Algorithm} \label{subsec:T2DRL_overview}
Generally, our T2DRL algorithm is based on the double deep Q-network (DDQN) algorithm~\cite{van2016deep} and the proposed diffusion-based deep deterministic policy gradient (D3PG) algorithm, each operating on different timescales. Specifically, within each time frame $t$, to handle the short-timescale resource allocation subproblem $\mathbf{P2}$, which is updated per time slot $k$, we propose the D3PG algorithm in Sec.~\ref{subsec:D3PG}. To address the long-timescale model caching subproblem $\mathbf{P3}$, which is updated per time frame $t$, the DDQN algorithm~\cite{van2016deep} is employed to adjust BS's cached GenAI models in Sec.~\ref{subsec:DDQN}. 


Finally, we propose a two-timescale deep reinforcement learning (T2DRL) algorithm that integrates the long-timescale DDQN algorithm with the short-timescale D3PG algorithm.

\vspace{-3mm}
\subsection{Short-Timescale Resource Allocation Using D3PG}\label{subsec:D3PG} 

To begin, we detail the motivation for adopting the DDPG framework, followed by a description of the MDP elements. Finally, we outline the architecture of the D3PG algorithm.

\subsubsection{Motivation of Adopting DDPG Framework}  \label{subsubsec:DDPG_motivation} 
Generally, DRL algorithms can be divided into two categories based on the form of the actions (i.e., variables to be solved): \emph{i) value-based DRL algorithms}, such as DQN~\cite{mnih2015human}, which are designed to handle discrete actions; \emph{ii) policy-based DRL algorithms}, with DDPG~\cite{lillicrap2015continuous} being a notable example, which focuses on solving continuous actions. Although value-based DRL algorithms can manage continuous action spaces by discretizing continuous actions into a set of potential values, coarse discretization results in a loss of behavioral detail, while overly fine discretization increases the dimensionality of the action space, making their convergence extremely slow.

As a result, we propose our D3PG algorithm based on the DDPG framework to directly optimize continuous variables, including the communication and computing resource allocation decisions, i.e., $b_{u,t}(k)$ and $\xi_{u,t}(k)$, respectively.

\subsubsection{MDP Elements in the D3PG} \label{subsubsec:D3PG_MDP}
The sequential decision-making nature of subproblem $\mathbf{P2}$ can be captured via an MDP, which includes the \emph{state space}, \emph{action space}, and \emph{reward function}, as described below.

\begin{itemize}[leftmargin=4mm]
\item \emph{State Space:} At time slot $k$ of time frame $t$, the DRL agent\footnote{The BS, which can gather information from all users, is defined as the central controller~\cite{liu2024dnn} and serves as the agent in both the D3PG and DDQN frameworks.} observes the state $\bm{s}_t(k)$ to gather environmental information, consisting of $4N+M$ elements, defined as follows:
\vspace{-1.5mm}
\begin{align}\label{eq:D3PG_state}
    \bm{s}_t(k)=\{\bm{h}_t(k),\bm{\varphi}_t(k),\bm{\varrho}(t), \bm{d}^{\mathsf{in}}_t(k), \bm{d}^{\mathsf{op}}_t(k)\},
    \vspace{-.5mm}
\end{align}
where $\bm{h}_t(k)=\{h_{u,t}(k)\}_{u\in\mathcal{U}}$ represents the channel gain vector for all users, $\bm{\varphi}_t(k)=\{\varphi_{u,t}(k)\}_{u\in\mathcal{U}}$ denotes the AIGC service request information for all users, $\bm{\varrho}(t)=\{\varrho_{m}(t)\}_{m\in\mathcal{M}}$ reflects the GenAI model caching decisions at current time frame $t$, $\bm{d}^{\mathsf{in}}_t(k)=\{d^{\mathsf{in}}_{u,t}(k)\}_{u\in\mathcal{U}}$ is the input data size vector for all users, and $\bm{d}^{\mathsf{op}}_t(k)=\{d^{\mathsf{op}}_{\varphi_{u,t}(k)}\}_{u\in\mathcal{U}}$ is the output data size vector of users' AIGC service requests.

\item \emph{Action Space:} At time slot $k$ of time frame $t$, the action space consists of communication and computing resource allocation decisions, containing $2N$ elements, expressed as
\vspace{-1.5mm}
\begin{align}\label{eq:D3PG_action}
    \bm{a}_t(k)=\{\bm{b}_t(k),\bm{\xi}_t(k)\},
    \vspace{-.5mm}
\end{align}
where $\bm{b}_t(k)=\{b_{u,t}(k)\}_{u\in\mathcal{U}}$ represents the bandwidth allocation ratio vector for all users, and $\bm{\xi}_t(k)=\{\xi_{u,t}(k)\}_{u\in\mathcal{U}}$ denotes the computing resource allocation ratio vector for all users. Note that the initial actions generated by the diffusion model are $\tilde{\bm{a}}_t(k)=\{\tilde{\bm{b}}_t(k),\tilde{\bm{\xi}}_t(k)\}$, with elements ranging from $[0,1]$. We then design an action amender~\cite{10418532} to ensure that all actions $\bm{a}_t(k)=\{\bm{b}_t(k),\bm{\xi}_t(k)\}$ satisfy the constraints outlined in subproblem $\mathbf{P2}$. Specifically, to satisfy constraint~\eqref{eq:Comm_Resource_Constraint}, the bandwidth allocation decision is modified to $b_{u,t}(k)=\frac{\tilde{b}_{u,t}(k)}{\sum_{u\in\mathcal{U}}\tilde{b}_{u,t}(k)}$. Similarly, to satisfy constraints~\eqref{eq:Comp_Resource_Constraint} and~\eqref{eq:Variable_Constraint}, the computing resource allocation decision is adjusted to $\xi_{u,t}(k)=\frac{\tilde{\xi}_{u,t}(k)\varrho_{m}(t)}{\sum_{u\in\mathcal{U}}\varrho_{m}(t)\tilde{\xi}_{u,t}(k)}$, where $m=\varphi_{u,t}(k)$.

\item \emph{Reward Function:} After executing action $\bm{a}_t(k)$ according to state $\bm{s}_t(k)$, the environment provides feedback in the form of a reward $r_t(k)$, which we define based on the objective function in~\eqref{eq:problem2} as follows:
\vspace{-1.5mm}
\begin{align}\label{eq:D3PG_reward}
    r_t(k)=-\frac{1}{U}\sum_{u\in\mathcal{U}} (G_{u,t}(k)+\mathbb{I}\{D^{\mathsf{tl}}_{u,t}>\tau\}\chi),
    \vspace{-.5mm}
\end{align}
where $\mathbb{I}\{\cdot\}$ is an indicator function with $\mathbb{I}\{\cdot\}=1$ when the condition is met; otherwise $\mathbb{I}\{\cdot\}=0$, and $\chi$ is a constant considered as the penalty to discourage the agent from violating constraint~\eqref{eq:Duration_Constraint}.
\end{itemize}

\subsubsection{Architecture of the D3PG Algorithm}  \label{subsubsec:D3PG_architecture}
The architecture of D3PG is depicted on the left side of Fig.~\ref{fig:T2DRL_Framework} and detailed below:


\begin{itemize}[leftmargin=4mm]
\item \emph{Diffusion Model-Based Actor Network:} In D3PG, the core of the actor network $\pi_{\bm{\theta}}$, parameterized by $\bm{\theta}$, is the denoiser in the diffusion model detailed in Sec.~\ref{sec:diffusion_model}, rather than a conventional MLP. Additionally, a target actor network, $\hat{\pi}_{\bm{\hat{\theta}}}$, parameterized by $\bm{\hat{\theta}}$, is employed to stabilize the learning process and has the same network structure as $\pi_{\bm{\theta}}$.

\item \emph{Critic Network:} The MLP-based critic network $Q_{\bm{\phi}}$, parameterized by $\bm{\phi}$, takes the state $\bm{s}_t(k)$ and action $\bm{a}_t(k)$ as inputs and outputs the Q-value function $Q_{\bm{\phi}}(\bm{s}_t(k),\bm{a}_t(k))$. Specifically, the Q-value function measures the quality of a state-action pair: a higher Q-value indicates that the corresponding state-action pair is likely to yield a higher reward. Similarly, a target critic network $\hat{Q}_{\bm{\hat{\phi}}}$, parameterized by $\bm{\hat{\phi}}$, is introduced to stabilize the learning process.

\item \emph{Replay Buffer:} During the training process, a replay buffer $\mathcal{E}$ is used to store the transition tuples. At each time slot $k$, D3PG stores $[\bm{s}_t(k), \bm{a}_t(k), r_t(k), \bm{s}_t(k+1)]$ in $\mathcal{E}$, where it awaits sampling for the development of network policies. 


\item \emph{Policy Improvement:} After a certain period of exploration, we randomly sample a mini-batch of $E$ samples $\{[\bm{s}^e_{t}(k),\bm{a}^e_{t}(k),r^e_{t}(k),\bm{s}^{e}_{t}(k+1)]\}_{e=1}^{E}$ from the replay buffer $\mathcal{E}$ to update the critic and actor networks. Specifically, for the critic network $Q_{\bm{\phi}}$, we minimize the average temporal difference error between the target Q-value $\hat{y}^e_{t}(k)$ and the Q-value $y^e_{t}(k)$, which is given by:
\vspace{-1.5mm}
\begin{align}\label{eq:D3PG_critic}
    &\quad\quad  \mathbb{L}^{\mathsf{D3PG}}=\frac{1}{E}\sum_{e=1}^E\Big[ \frac{1}{2}(\hat{y}^e_{t}(k)-y^e_{t}(k))^2  \Big], \\
    &\hspace{-3mm}\text{s.t.}~\ y^e_{t}(k)=Q_{\bm{\phi}}(\bm{s}^e_{t}(k),\bm{a}^e_{t}(k)) \tag{24a},\\
    & \hat{y}^e_{t}(k)=r^e_{t}(k)+\omega \hat{Q}_{\bm{\hat{\phi}}}(\bm{s}^{e}_{t}(k+1),\hat{\pi}_{\hat{\bm{\theta}}}(\bm{s}^{e}_{t}(k+1))). \tag{24b}
    \vspace{-.5mm}
\end{align}
Here, $e$ represents the $e$-th transition tuple sampled from the replay buffer $\mathcal{E}$, and $\omega$ denotes the discount factor for future rewards in D3PG. The target Q-value $\hat{y}^e_{t}(k)$ is calculated through the target critic network $\hat{Q}_{\bm{\hat{\phi}}}$. Specifically, the target critic network takes the next state $\bm{s}^{e}_{t}(k+1)$ and the next action $\hat{\pi}_{\hat{\bm{\theta}}}(\bm{s}^{e}_{t}(k+1))$ generated by the target actor network as inputs and outputs the corresponding target Q-value. The estimation accuracy of $Q_{\bm{\phi}}$ is then improved by iteratively minimizing the loss in~\eqref{eq:D3PG_critic} using a standard optimizer, such as Adam~\cite{kingma2014adam} as
\vspace{-1.5mm}
\begin{align}\label{eq:D3PG_criticnet_update}
    \bm{\phi} \leftarrow \bm{\phi} - \sigma^{\mathsf{c}}\mathbb{L}^{\mathsf{D3PG}},
    \vspace{-.5mm}
\end{align}
where $\sigma^{\mathsf{c}}$ represents the learning rate of the critic network. On the other hand, the actor network $\pi_{\bm{\theta}}$ is updated using the sample policy gradient:
\vspace{-1.5mm}
\begin{align}\label{eq:D3PG_actor}
    &\nabla_{\pi_{\bm{\theta}}}\mathcal{J}= \nonumber \\ 
    &\frac{1}{E}\sum_{e=1}^E  \nabla_{\bm{a}}Q_{\bm{\phi}}(\bm{s}^e_{t}(k),\bm{a})_{\bm{a}=\pi_{\bm{\theta}}(\bm{s}^e_{t}(k))}\nabla_{\bm{\theta}}\pi_{\bm{\theta}}(\bm{s}^e_{t}(k)),
    \vspace{-1.5mm}
\end{align}
where the actor network $\pi_{\bm{\theta}}$ is updated by gradient ascent iterations using~\eqref{eq:D3PG_actor} to maximize the reward in~\eqref{eq:D3PG_reward}, e.g., using Adam~\cite{kingma2014adam}, as follows: 
\vspace{-1.5mm}
\begin{align}\label{eq:D3PG_actornet_update}
    \bm{\theta} \leftarrow \bm{\theta} + \sigma^{\mathsf{a}}\nabla_{\pi_{\bm{\theta}}}\mathcal{J},
    \vspace{-.5mm}
\end{align}
where $\sigma^{\mathsf{a}}$ represents the learning rate of the actor network. During the training process, the parameters of the target networks are updated slowly to ensure smooth changes in the policy and Q-value function over time, as described mathematically:
\vspace{-1.5mm}
\begin{align}
    &\hat{\bm{\theta}}\leftarrow \varepsilon\bm{\theta}+(1-\varepsilon)\hat{\bm{\theta}}, \label{eq:D3PG_target_actor}\\
    &\hat{\bm{\phi}} \leftarrow \varepsilon\bm{\phi}+(1-\varepsilon)\hat{\bm{\phi}}, \label{eq:D3PG_target_critic}
    \vspace{-.5mm}
\end{align}
where $\varepsilon \in (0,1]$ is the target network update rate in D3PG.

\end{itemize}

\vspace{-4mm}
\subsection{Long-Timescale Model Caching Using DDQN}\label{subsec:DDQN} 
Similarly, we first outline the motivation for using DDQN, then describe the corresponding elements in its MDP. Finally, we detail the architecture of the DDQN algorithm.

\subsubsection{Motivation of Utilizing DDQN Algorithm} Recalling that the GenAI model caching decision $\varrho_m(t)\in \{0,1\}$ is binary, thus the corresponding action space of subproblem $\mathbf{P3}$ is discrete. Consequently, DQN, the value-based DRL algorithm, can obtain the model caching decisions. However, in DQN, the same Q-network is used to both select and evaluate actions, which can lead to overestimation~\cite{10497174}, especially when the estimates are based on an inaccurate model of the environment.

To address this, we utilize the double deep Q-network (DDQN)~\cite{van2016deep} to separate action selection from action evaluation. Specifically, DDQN uses an evaluation Q-network to select the best action and a target Q-network to determine its Q-value (details will be provided in Sec.~\ref{subsubsec:DDQN_architecture}), resulting in more accurate value estimates and improved performance.

\subsubsection{MDP Elements in the DDQN} \label{subsubsec:DDQN_MDP} We next describe the \emph{state space}, \emph{action space}, and \emph{reward function} within the MDP for subproblem $\mathbf{P3}$.

\begin{itemize}[leftmargin=4mm]
\item \emph{State Space:} At time frame $t$, the system state $s(t)$ is defined to contain the corresponding skewness of popularity as
\vspace{-1.5mm}
\begin{align}\label{eq:DDQN_state}
    s(t)=\{\gamma(t)\}.
    \vspace{-.5mm}
\end{align}

\item \emph{Action Space:} At time frame $t$, the action space consists of GenAI model caching decisions, containing $M$ elements, expressed as:
\vspace{-1.5mm}
\begin{align}\label{eq:DDQN_action}
    \bm{a}(t) = \{\bm{\varrho}(t)\},
    \vspace{-.5mm}
\end{align}
where $\bm{\varrho}(t)=\{\varrho_m(t)\}_{m\in \mathcal{M}}$ is the caching decision vector for all GenAI models. Also, given the initial action generated by the agent $\tilde{a}(t) \in \{0,1,...,2^{M}-1\}$, we  use an action amender to satisfy~\eqref{eq:Caching_Constraint}, where $\varrho_{m}(t)= \lfloor\frac{\tilde{a}(t)}{2^{M-m}}\rfloor$ mod 2.


\item \emph{Reward Function:} At the beginning of each time frame $t$, after the BS takes action $\bm{a}(t)$, the corresponding model caching decisions remain unchanged for the following $K$ time slots. Once this time frame ends, the time slot rewards $r_t(k)$ given in~\eqref{eq:D3PG_reward} for the $K$ time slots within time frame $t$ will be captured. Consequently, the time frame reward $r(t)$ is calculated as the average of these feedback values, aligned with the negative value of~\eqref{eq:problem3} as follows:
\vspace{-1.5mm}
\begin{align}\label{eq:DDQN_reward}
    r(t)=-\frac{1}{K}\sum_{k\in\mathcal{K}} r_t(k) -\mathbb{I}\{\sum_{m \in \mathcal{M}}\varrho_m(t)c_m > C\}\Xi,
    \vspace{-.5mm}
\end{align}
where $\Xi$ is a constant penalty used to prevent the agent from violating constraint~\eqref{eq:Caching_Resource_Constraint}.
\end{itemize}

\begin{algorithm} [!t]
  \SetAlgoLined
  \SetKwData{Left}{left}\SetKwData{This}{this}\SetKwData{Up}{up}
  \SetKwFunction{Union}{Union}\SetKwFunction{FindCompress}{FindCompress}
  \SetKwInOut{Input}{input}\SetKwInOut{Output}{output}
{\footnotesize
  \textbf{Input:} Initialize the network parameters $\bm{\theta}$, $\bm{\phi}$, and $\bm{\eta}$, and set values for all hyperparameters, including the learning episode $H$, time frame length $T$, time slot length $K$, discount factors $\omega$ and $\rho$, and penalties $\chi$ and $\Xi$, among others.
  
  
  \textbf{Output:} Well-trained network parameters: $\bm{\theta}$ and $\bm{\eta}$.
  \BlankLine
  
  \For{$episode=1$ \KwTo $H$}{
    \textbf{\textit{// Start of DDQN algorithm //}}
    
    \For{$t=1$ \KwTo $T$}{
        Observe the environment to obtain $s(t)$ according to~\eqref{eq:DDQN_state} and get the action $\bm{a}(t) = \{\bm{\varrho}(t)\}$.
        
        The corresponding model caching decisions $\bm{\varrho}(t)$ are obtained and remain unchanged for the upcoming $K$ time slots.

        \textbf{\textit{// Start of D3PG algorithm //}}
        
        \For{$k=1$ \KwTo $K$}{ 
            Observe the environment to obtain $\bm{s}_t(k)$ according to~\eqref{eq:D3PG_state} and initialize a distribution $\bm{x}^L_t(k)\sim\mathcal{N}(0,\mathbf{I})$.
      
            \For{$l=L$ \KwTo $0$} {
            Use a deep neural network to infer the noise $\hat{\bm{\epsilon}}_{\bm{\theta}}(\bm{x}^l_t(k),l,\bm{s}_t(k))$.
        
            Calculate the mean $\bm{\mu}_{\bm{\theta}}(\bm{x}^{l}_t(k),l,\bm{s}_t(k))$ and the distribution $q(\bm{x}^{l-1}_t(k)|\bm{x}^{l}_t(k))$ by~\eqref{eq:reverse_mean} and~\eqref{eq:reverse_distribution}, respectively. 

            Calculate the distribution $\bm{x}^{l-1}_t(k)$ using the reparameterization technique~\eqref{eq:reverse_update}.
		  }
      
        Obtain the optimal resource allocation decisions $\bm{x}^0_t(k)=\{\bm{b}^*_t(k),\bm{\xi}^*_t(k)\}$.
        
        Receive the reward $r_t(k)$ according to~\eqref{eq:D3PG_reward} and transition to the next state $\bm{s}_t(k+1)$.
      
        Store $[\bm{s}_t(k),\bm{a}_t(k),r_t(k),\bm{s}_{t}(k+1)]$ into $\mathcal{E}$. 
        
        Randomly sample a batch of $E$ transitions $\{[\bm{s}^e_{t}(k),\bm{a}^e{t}(k),r^e_{t}(k),\bm{s}^{e}_{t}(k+1)]\}_{e=1}^E$ from $\mathcal{E}$.

        Update networks' parameters $\bm{\phi}$ and $\bm{\theta}$ by minimizing ~\eqref{eq:D3PG_criticnet_update} and~\eqref{eq:D3PG_actornet_update}, respectively.

        Update the target networks' parameters $\hat{\bm{\theta}}$ and $\bm{\hat{\phi}}$ by~\eqref{eq:D3PG_target_actor} and~\eqref{eq:D3PG_target_critic}, respectively.
        }
      \textbf{\textit{// End of D3PG algorithm //}}
    
      Receive the time frame reward $r(t)$ according to~\eqref{eq:DDQN_reward} based on the average of $K$ time slot rewards $r_t(k)$ and transition to the next state $s(t+1)$.   
 
      Store $[s(t),a(t),r(t),s(t+1)]$ into $\mathcal{F}$.
      
      Randomly sample a batch of $F$ transitions $\{[s^f(t),a^f(t),r^f(t),s^{f}(t+1)]\}_{f=1}^F$ from $\mathcal{F}$.
      
      Update the evaluation Q-network's parameters $\bm{\eta}$ by minimizing~\eqref{eq:DDQN_criticnet_update}.
      
      Update the target Q-network's parameters $\bm{\hat{\eta}}$ by~\eqref{eq:DDQN_target_critic}.
    }   
    \textbf{\textit{// End of DDQN algorithm //}}
  }}
  \caption{T2DRL Algorithm}\label{algo：T2DRL}
\end{algorithm}\DecMargin{1em}

\subsubsection{Architecture of the DDQN Algorithm} \label{subsubsec:DDQN_architecture}
The architecture of DDQN is depicted on the right side of Fig.~\ref{fig:T2DRL_Framework}, detailed below:


\begin{itemize}[leftmargin=4mm]
\item \emph{Evaluation Q-Network:} At each time frame $t$, the MLP-based evaluation Q-network, parameterized by $\bm{\eta}$, takes the state $s(t)$ and action $\bm{a}(t)$ as inputs and evaluates the corresponding Q-value $Q_{\bm{\eta}}(s(t), \bm{a}(t))$, with a higher Q-value indicating a better state-action pair.  

\item \emph{Target Q-Network:} The MLP-based target Q-network, parameterized by $\bm{\hat{\eta}}$, takes the next state $s(t+1)$ and the next action $\bm{a}(t+1) = \text{argmax}_{a}Q_{\bm{\eta}}(s(t+1), a)$, produced by the evaluation Q-network, as inputs and then outputs the corresponding target Q-value $\hat{Q}_{\bm{\hat{\eta}}}(s(t+1), \bm{a}(t+1))$.

\item \emph{Replay Buffer:} DDQN utilizes a replay buffer $\mathcal{F}$ to store the transition tuple $[s(t), \bm{a}(t), r(t), s(t+1)]$, which is then sampled to improve the accuracy of evaluation Q-network.

\item \emph{Policy Improvement:} A mini-batch of $F$ data samples $\{[s^f(t),\bm{a}^f(t),r^f(t),s^{f}(t+1)]\}_{f=1}^{F}$ is randomly sampled from the replay buffer $\mathcal{F}$ to update the evaluation Q-network. Specifically, this update is done to minimize the average temporal difference error between the target Q-value $\hat{y}^f(t)$ and the evaluation Q-value $y^f(t)$, which is given by
\vspace{-1.5mm}
\begin{align}\label{eq:DDQN_critic}
    &\quad\quad \mathbb{L}^{\mathsf{DDQN}}= \frac{1}{F}\sum_{f=1}^F\Big[ \frac{1}{2}(\hat{y}^f(t)-y^f(t))^2  \Big], \\
    &\text{s.t.} \quad \hat{y}^f(t)= \nonumber\\
    &r^f(t)+\rho \hat{Q}_{\bm{\hat{\eta}}}(s^{f}(t+1),\text{argmax}_{a}Q_{\bm{\eta}}(s^{f}(t+1), a)), \tag{33a} \\
    & y^f(t) = Q_{\bm{\eta}}(s^f(t),\bm{a}^f(t)) \tag{33b}, 
    \vspace{-.5mm}
\end{align}
where $f$ represents the $f$-th transition tuple sampled from the replay buffer $\mathcal{F}$, and $\rho$ represents the discount factor for future rewards in DDQN. The estimation accuracy of $Q_{\bm{\eta}}$ is then improved by iteratively minimizing the loss in~\eqref{eq:DDQN_critic} using a standard optimizer, such as Adam~\cite{kingma2014adam}, as given by
\vspace{-1.5mm}
\begin{align}\label{eq:DDQN_criticnet_update}
    \bm{\eta} \leftarrow \bm{\eta} - \zeta\mathbb{L}^{\mathsf{DDQN}},
    \vspace{-.5mm}
\end{align}
where $\zeta$ is the learning rate of the evaluate network. Similarly, the parameters of the target Q-network are slowly updated towards the evaluation Q-network as follows:
\vspace{-1.5mm}
\begin{align}
    &\hat{\bm{\eta}} \leftarrow \kappa\bm{\eta}+(1-\kappa)\hat{\bm{\eta}}, \label{eq:DDQN_target_critic}
    \vspace{-.5mm}
\end{align}
where $\kappa \in (0,1]$ is the target network update rate in DDQN.

\end{itemize}

\subsection{T2DRL Algorithm and Complexity Analysis}
Algorithm~\ref{algo：T2DRL} outlines the pseudocode of our proposed T2DRL, which consists of the outer DDQN algorithm in the long-timescale (\emph{lines 4-30}) and the inner D3PG algorithm in the short-timescale (\emph{lines 8-23}). Since the training process of learning-based methods can be implemented on a cloud data center with sufficient computational resources (e.g., as in~\cite{liu2024dnn}), we primarily focus on the computational complexity of the execution process (i.e., inference process) of the DRL.

First, we analyze the computational
complexity of the inference process in the DDQN algorithm. Assuming the DDQN contains $L^{\mathsf{q}}$ fully connected layers, each with $n^{\mathsf{q}}_l$ neurons, the computational complexity for each time frame $t$ can be denoted as $\mathcal{O}(|s(t)|n^{\mathsf{q}}_1+\sum_{l=1}^{L^{\mathsf{q}}-1}n^{\mathsf{q}}_{l}n^{\mathsf{q}}_{l+1}+n^{\mathsf{q}}_{L^{\mathsf{q}}}2^M)$. Similarly, we analyze the computational complexity of the inference process in the D3PG algorithm. Assuming the D3PG contains $L^{\mathsf{d}}$ fully connected layers, each with $n^{\mathsf{d}}_l$ neurons, the computational complexity for each time slot $k$ with $L$ denoising steps can be expressed as $\mathcal{O}(L(|\bm{s}_t(k)|n^{\mathsf{d}}_1+\sum_{l=1}^{L^{\mathsf{d}}-1}n^{\mathsf{d}}_{l}n^{\mathsf{d}}_{l+1}+n^{\mathsf{d}}_{L^{\mathsf{d}}}2U))$.

To sum up, for $H$ episodes, $T$ time frames, $K$ time slots, and $L$ denoising steps in the diffusion model, the overall computational complexity of the inference process for the proposed T2DRL algorithm is $\mathcal{O}(HT[(|s(t)|n^{\mathsf{q}}_1+\sum_{l=1}^{L^{\mathsf{q}}-1}n^{\mathsf{q}}_{l}n^{\mathsf{q}}_{l+1}+n^{\mathsf{q}}_{L^{\mathsf{q}}}2^M)KL(|\bm{s}_t(k)|n^{\mathsf{d}}_1+\sum_{l=1}^{L^{\mathsf{d}}-1}n^{\mathsf{d}}_{l}n^{\mathsf{d}}_{l+1}+n^{\mathsf{d}}_{L^{\mathsf{d}}}2U)])$.

\vspace{-2mm}
\section{Performance Evaluation}  \label{sec:simulation}  
In this section, we first outline the parameter settings for simulations and then evaluate the performance of our proposed T2DRL by comparing it against three benchmark methods.

\begin{table}[h]
\vspace{-2mm}
\centering
\footnotesize
\caption{Parameters used in simulations ~\cite{zhang2022two, park2020mobile, yao2023cooperative}.}
\label{table1}
\rowcolors{1}{lightblue!30}{white}
\begin{tabular}{|l|c|}
\hline
\textbf{Parameter}                    & \textbf{Value}   \\ \hline \hline
Number of time frames $T$                  & 10                \\ \hline
Number of time slots $K$                   & 10                \\ \hline
Duration of each time slot $\tau$          & 20 seconds        \\ \hline
Input data size $d^{\mathsf{in}}_{u,t}(k)$   & [5, 10] MB           \\ \hline
Uplink bandwidth $W^{\mathsf{up}}$                  & 20 MHz                \\ \hline
Downlink bandwidth $W^{\mathsf{dw}}$                  & 40 MHz                \\ \hline
Transmit power of user $p_u$                  & 23 dBm        \\ \hline
Transmit power of the BS $p^{\mathsf{BS}}$                  & 43 dBm        \\ \hline
Noise power spectral density $N_0$                  & -176 dBm/Hz        \\ \hline
Backhual transmission rate $R^{\mathsf{bc}}$, $R^{\mathsf{cb}}$                  & 100 Mbps          \\ \hline
Output data size $d^{\mathsf{op}}_{m}$              & [5, 10] MB           \\ \hline
Number of denoising steps performed at the BS $\mathcal{L}$         & 1000                \\ \hline
Storage capacity of the BS $C$    & 20 GB             \\ \hline
Preference weight factor $\alpha$  & 0.7              \\ \hline
D3PG's reward penalty $\chi$                  & 10          \\ \hline
D3PG's target network update rate $\varepsilon$                          & 0.005        \\ \hline
DDQN's reward penalty $\Xi$                  & 100          \\ \hline
DDQN's target network update rate $\kappa$                          & 0.005        \\\hline 
Number of episodes $H$                          & 500        \\\hline
\end{tabular}
\end{table}

\subsection{Simulation Settings}
\emph{1) Network Layout:} We consider a network consisting of a cloud data center, a BS (i.e., an edge server), and $[10, 18]$ users within a 250 m $\times$ 250 m square area. User mobility is modeled as a finite Markov state transition model with $I=3$ states: ${\lambda_1,\lambda_2,\lambda_3}$, representing uniform, concentrated, and boundary distributions of user locations, respectively. We assume that the transition probabilities between the user location distribution states are as follows:
\begin{equation} \label{eq:user_location_states}
P^{\Lambda} = 
\begin{bmatrix}
P_{11}^{\Lambda} & P_{12}^{\Lambda} & P_{13}^{\Lambda} \\
P_{21}^{\Lambda} & P_{22}^{\Lambda} & P_{23}^{\Lambda} \\
P_{31}^{\Lambda} & P_{32}^{\Lambda} & P_{33}^{\Lambda} \\
\end{bmatrix}
=
\begin{bmatrix}
0.6 & 0.1 & 0.3 \\
0.3 & 0.6 & 0.1 \\
0.1 & 0.3 & 0.6 \\
\end{bmatrix}.
\end{equation}

\emph{2) GenAI Models:} We consider that the number of model types is $M=10$, with the storage size of each GenAI model $c_m \in [2, 10]$ GB. To simulate the varying capabilities of different GenAI models, we set $A_1 \in [50,100]$, $A_2 \in [100,150]$, $A_3 \in [150,200]$, $A_4 \in (0,50]$, $B_1 \in (0,0.5]$, and $B_2 \in (0,10]$. Similarly, we model the AIGC service popularity as a finite Markov state transition model with $J=3$ states: ${\gamma_1,\gamma_2,\gamma_3}$, which have parameters $\gamma_1=0.2, \gamma_2 = 0.5$, and $\gamma_3=0.7$, respectively. We then assume that the transition probabilities between the AIGC service popularity states are as follows:
\begin{equation} \label{eq:AIGC_service_states}
P^{\Gamma} = 
\begin{bmatrix}
P_{11}^{\Gamma} & P_{12}^{\Gamma} & P_{13}^{\Gamma} \\
P_{21}^{\Gamma} & P_{22}^{\Gamma} & P_{23}^{\Gamma} \\
P_{31}^{\Gamma} & P_{32}^{\Gamma} & P_{33}^{\Gamma} \\
\end{bmatrix}
=
\begin{bmatrix}
0.6 & 0.2 & 0.2 \\
0.1 & 0.7 & 0.2 \\
0.2 & 0.3 & 0.5 \\
\end{bmatrix}.
\end{equation}

\emph{3) Experimental Platform:} We implement T2DRL using PyTorch 2.0 and Python 3.8.1 on a platform with an NVIDIA RTX A5000 GPU. For the diffusion model, we configure the DNNs to learn noise using 3 fully connected (FC) hidden layers, each consisting of 128 neurons. The D3PG critic networks are built with 2 FC hidden layers, each containing 256 neurons. Similarly, the Q-networks of the DDQN are constructed with 2 FC hidden layers, each containing 128 neurons. We use Adam optimizer with a learning rate of $1 \times 10^{-6}$ for both D3PG and DDQN, and we apply ReLU as the activation function for each hidden layer. The detailed settings for other parameters in our experiments are presented in Table~\ref{table1}.

\vspace{-1.5mm}
\subsection{Benchmark Solutions}
To demonstrate the effectiveness of the proposed T2DRL algorithm, we have relied on three benchmark solutions:

\begin{itemize}[leftmargin=4.5mm]
	\item \emph{DDPG-based T2DRL:} The long-timescale model caching placement is optimized using the same policy as in T2DRL (i.e., DDQN), while the short-timescale resource allocation decision is updated by the DDPG algorithm~\cite{lillicrap2015continuous}. In contrast to our proposed D3PG, DDPG uses an MLP-based actor network to make optimal resource allocation decisions. This method is used to highlight the significant advantages of the diffusion model leveraged in this work.
    \item \emph{Static caching and heuristic-based resource allocation scheme (SCHRS):} Drawing inspiration from~\cite{10609797}, the BS caches the most popular GenAI models based on a popularity skewness of $\gamma_1=0.2$ without considering the dynamic AIGC service popularity, while bandwidth and computing resource allocation decisions are generated by a genetic algorithm. Specifically, SCHRS employs real-valued encoding to generate multiple chromosomes (solutions), forming the initial population. These chromosomes are then evolved through a series of simulated binary crossovers and polynomial mutations until the maximum number of iterations is reached. Finally, the chromosome with the lowest value of~\eqref{eq:problem2} is selected.
    \item \emph{Randomized caching and average resource allocation scheme (RCARS)~\cite{nath2020deep}:} At each time frame, the BS randomly caches GenAI models until reaching the caching capacity, while bandwidth and computing resources are evenly distributed among users at each time slot. This approach serves as a lower-bound baseline for evaluating the performance.
    
\end{itemize}

\vspace{-3mm}
\subsection{Simulation Results}


\subsubsection{Convergence Performance}
\begin{figure}[t!]
    \centering
    \begin{subfigure}[b]{.38\textwidth}
        \includegraphics[width=\textwidth]{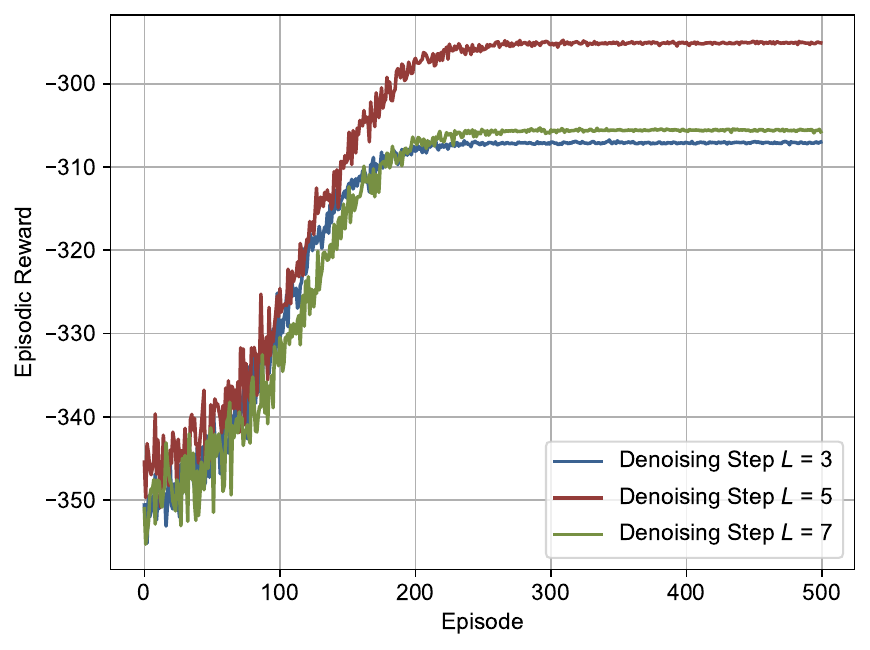}
        \vspace{-7mm}
        \caption{Denoising step impact on the reward.}
        \label{fig:T2DRL_convergency_different_steps}
    \end{subfigure}
    \begin{subfigure}[b]{.38\textwidth}
        \includegraphics[width=\textwidth]{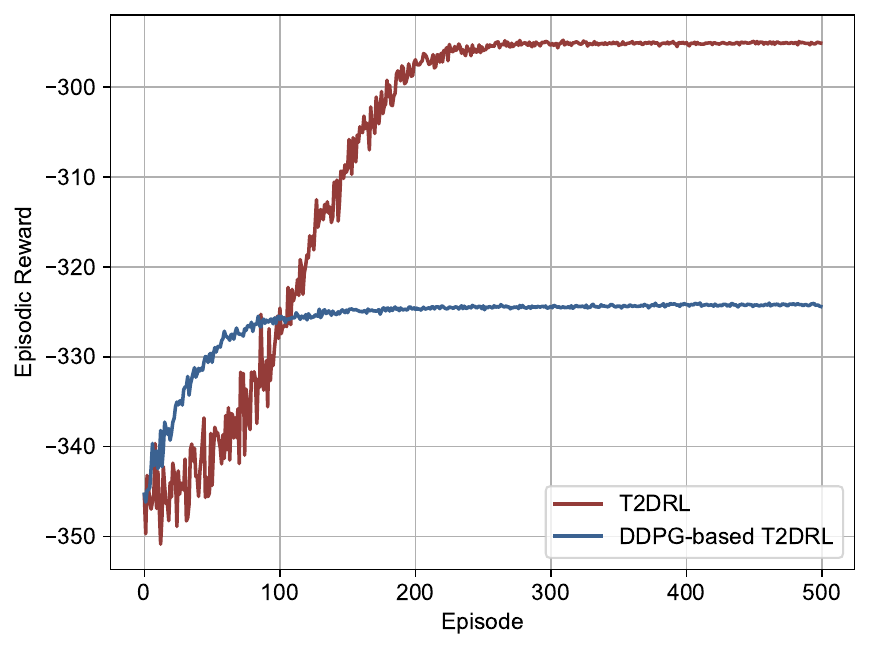}
        \vspace{-7mm}
        \caption{Comparison of reward curves of T2DRL and DDPG-based T2DRL.}
        \label{fig:D3PG_DDPG_convergency}
    \end{subfigure}
    \vspace{-0.5mm}
    \caption{Convergence performance analysis.}
    \label{fig:test}
\end{figure}

Since the number of denoising steps $L$ in the reverse process guides the action sampling procedure, we present the episodic reward curves for T2DRL under various denoising step settings. As shown in Fig.~\ref{fig:T2DRL_convergency_different_steps}, we observe that the converged reward initially increases but then decreases as the number of denoising steps increases. The initial increase in the reward occurs because, with more denoising steps, the training process becomes more stable, allowing the diffusion model to learn more general features. However, as the number of denoising steps continues to increase, the diffusion model may remove too much noise, stripping valuable details from the data and resulting in a worse performance. Therefore, we set the denoising steps of D3PG $L=5$ for comparison with benchmark solutions in the subsequent results.

In Fig.~\ref{fig:D3PG_DDPG_convergency}, we depict the convergence behavior of both our proposed T2DRL and the DDPG-based T2DRL as the number of training episodes increases. Overall, our proposed T2DRL achieves a higher episodic reward, highlighting the substantial benefits of the diffusion model. This improvement is largely attributed to the generative capabilities of the diffusion model, which enhance action sampling efficiency by progressively reducing noise through the reverse process. 


\subsubsection{Effect of the Number of Users}
\begin{figure*}[t!]
    \centering
    \begin{subfigure}[b]{.38\textwidth}
        \includegraphics[width=\textwidth]{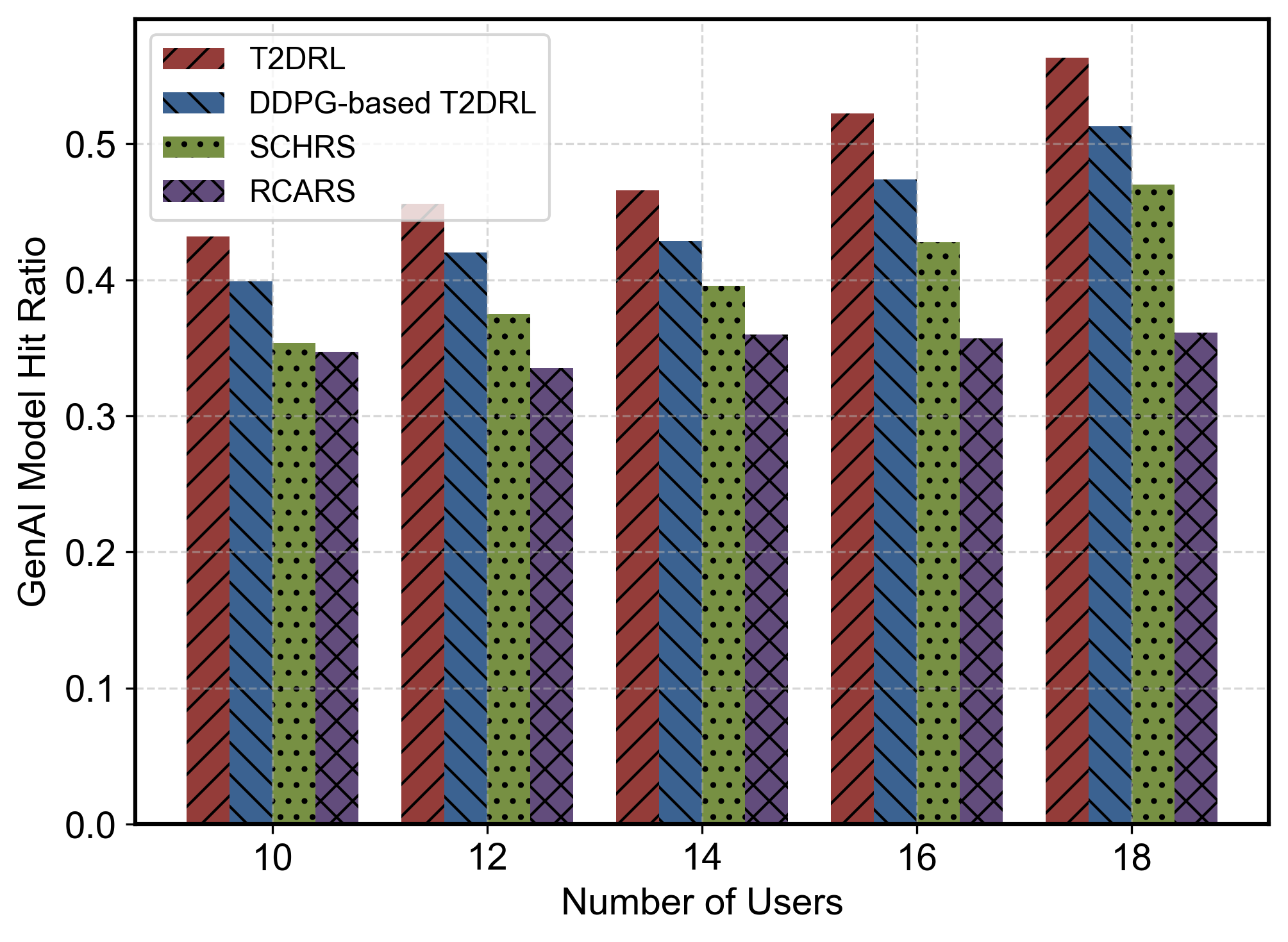}
        \vspace{-5mm}
        \caption{Impact on the GenAI model hit ratio.}
        \label{fig:Model_hit_ratio_different_users}
    \end{subfigure}
    \begin{subfigure}[b]{.38\textwidth}
        \includegraphics[width=\textwidth]{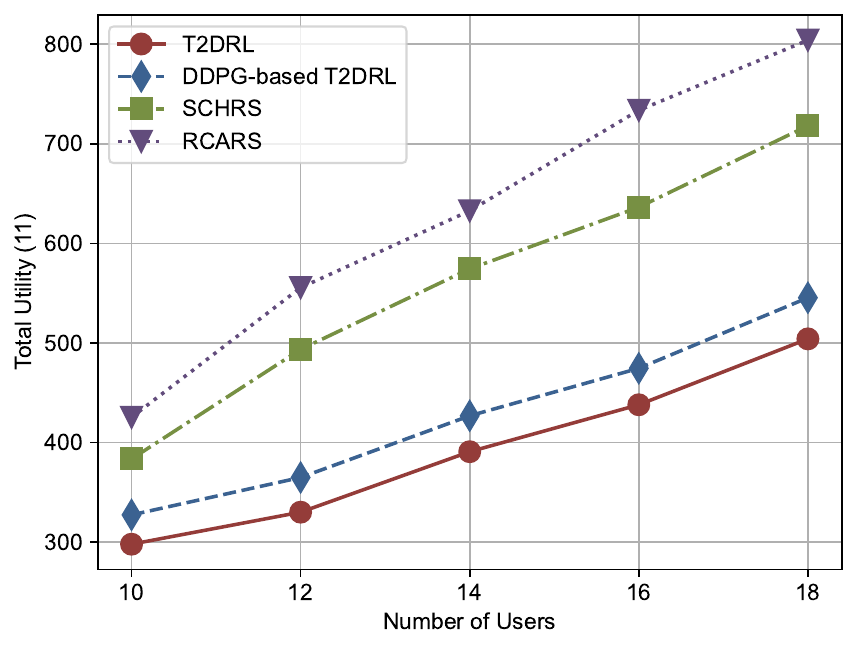}
        \vspace{-7mm}
        \caption{Impact on the total utility~\eqref{eq:problem1}.}
        \label{fig:Objective_value_different_users}
    \end{subfigure}
    \vspace{-0.5mm}
    \caption{Performance evaluations upon considering different numbers of users.}
    \label{fig:test}
\end{figure*}

The results shown in Fig.~\ref{fig:Model_hit_ratio_different_users} illustrate the effect of incrementally increasing the number of users from 10 to 18 on the GenAI model hit ratio. We observe that as the number of users increases, the model hit ratio improves. This is because the AIGC service requests begin to more closely reflect the underlying probability distribution, with popular AIGC services accounting for a larger proportion of total requests. Overall, our proposed T2DRL outperforms the other algorithms. Its performance is 8.28\% better than the DDPG-based T2DRL, 22.06\% better than SCHRS, and 24.43\% better than RCARS at 10 users. Also, its performance is 9.81\% better than the DDPG-based T2DRL, 21.86\% better than SCHRS, and 56.01\% better than RCARS at 18 users.

Fig.\ref{fig:Objective_value_different_users} illustrates the effect of the number of users on the total utility~\eqref{eq:problem1}. We observe that as the number of users increases, the total utility~\eqref{eq:problem1} also rises. This is because, with a growing number of users and unchanged available bandwidth and computing resources, competition among users intensifies, leading to fewer resources available for each user. Overall, T2DRL outperforms the other algorithms. Its performance is 8.98\% better than the DDPG-based T2DRL, 22.31\% better than SCHRS, and 29.95\% better than RCARS at 10 users; and 9.57\% better than the DDPG-based T2DRL, 29.79\% better than SCHRS, and 37.29\% better than RCARS at 18 users.

\begin{figure*}[t!]
    \centering
    \begin{subfigure}[b]{.39\textwidth}
        \includegraphics[width=\textwidth]{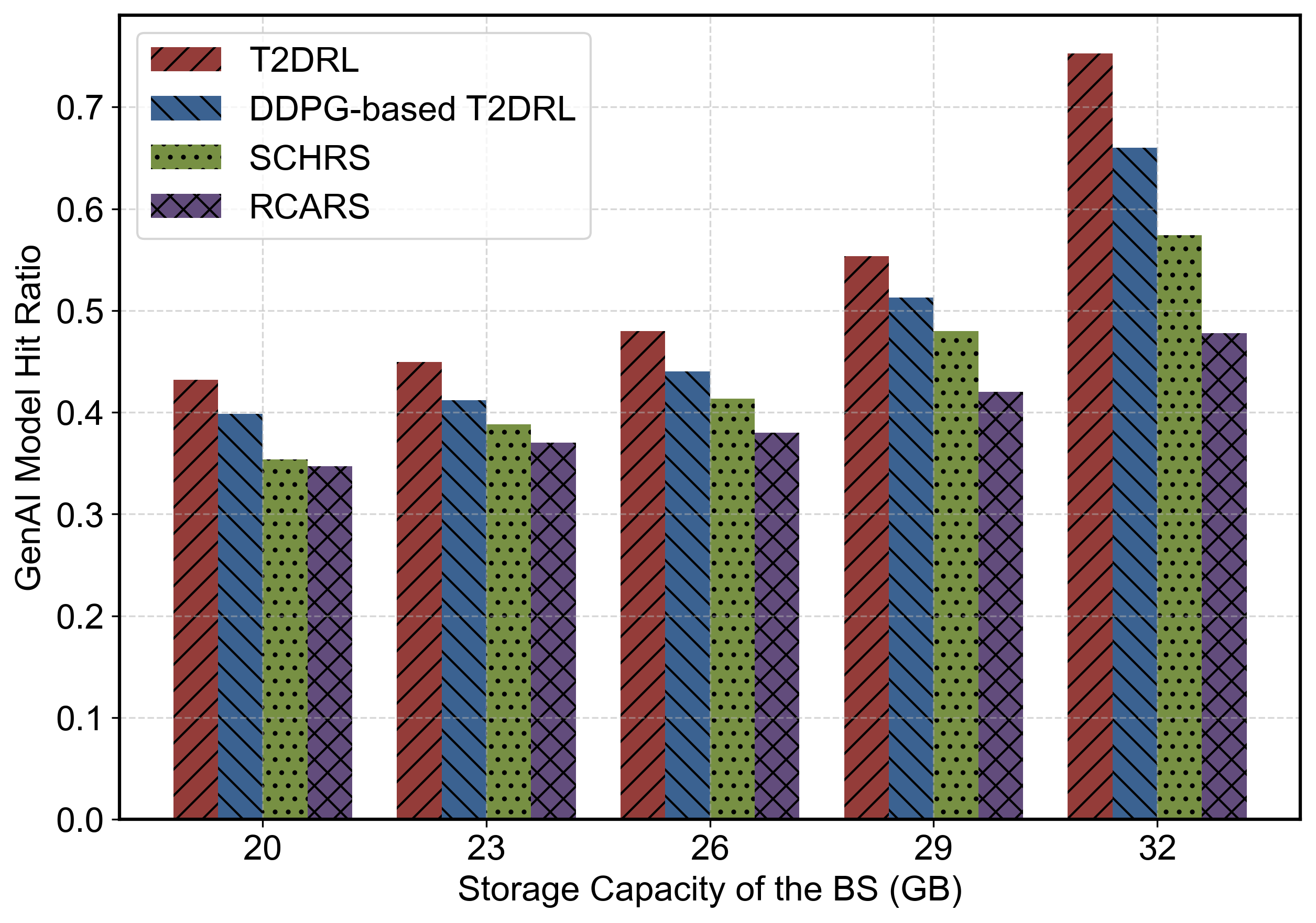}
        \vspace{-5mm}
        \caption{Impact on the GenAI model hit ratio.}
        \label{fig:Model_hit_ratio_different_caching}
    \end{subfigure}
    \begin{subfigure}[b]{.38\textwidth}
        \includegraphics[width=\textwidth]{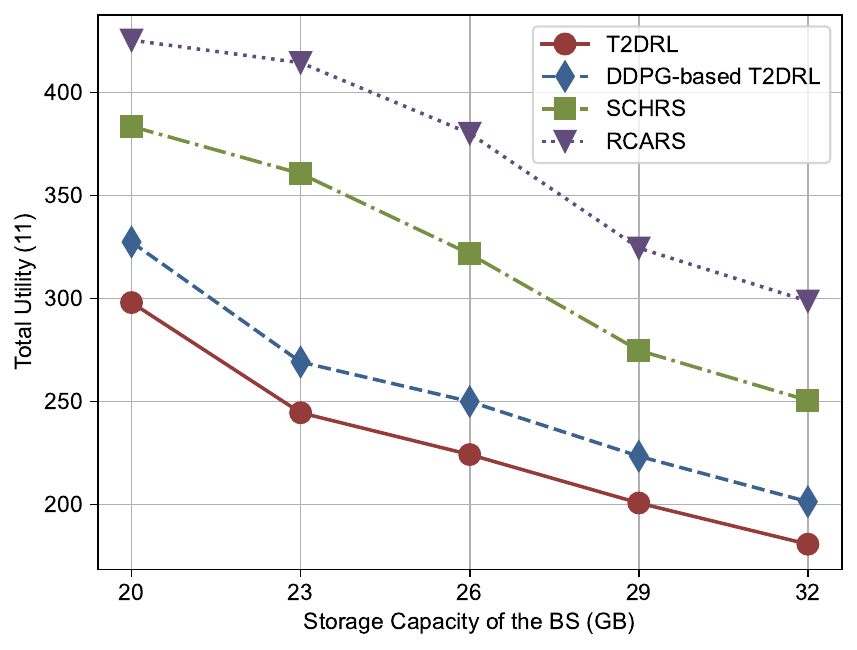}
        \vspace{-7mm}
        \caption{Impact on the total utility~\eqref{eq:problem1}.}
        \label{fig:Objective_value_different_caching}
    \end{subfigure}
    \vspace{-0.5mm}
    \caption{Performance evaluations upon considering different storage capacities of the BS.}
    \label{fig:test}
\end{figure*}

\begin{table}[htbp]
\centering
\footnotesize
\caption{Comparison of algorithm running time per time slot (millisecond).}
\label{table2}
\rowcolors{1}{lightblue!30}{white}
\begin{tabular}{|c|c|c|c|c|c|}
\hline
\textbf{Number of Users} & \textbf{10} & \textbf{12} & \textbf{14} & \textbf{16} & \textbf{18} \\ \hline\hline
T2DRL & 1.112 & 1.162 & 1.197 & 1.226 & 1.272\\ \hline
DDPG-based T2DRL & 0.302 & 0.317 & 0.324 & 0.339 & 0.351 \\ \hline
SCHRS & 624.9 & 769.7 & 844.1 & 973.2 & 1128.7 \\ \hline
\end{tabular}
\end{table}

Table~\ref{table2} presents the effect of the number of users on the algorithm's running time per time slot. RCARS is excluded from the comparisons due to its poor performance in terms of the total utility~\eqref{eq:problem1}. We observe that SCHRS has the longest running time, as evolving the initial chromosomes requires extensive generations to produce relatively good offspring. On the other hand, the running time of the proposed T2DRL is longer than that of the DDPG-based T2DRL, primarily due to the additional reverse process with the denoising step $L=5$. However, given T2DRL's highest model hit ratio and lowest total utility~\eqref{eq:problem1}, we can conclude that T2DRL performs well, with a slight increase in computation complexity.







\subsubsection{Effect of the Caching Space of the Edge Server}
Fig.~\ref{fig:Model_hit_ratio_different_caching} illustrates the effect of the edge server's caching capacity on the GenAI model hit ratio. We observe that as the edge server's caching space increases, the model hit ratio for all algorithms steadily improves. This is because a larger cache allows the edge server to store more GenAI models, enabling user requests to be fulfilled directly from the cache rather than fetching them from the cloud. Overall, our proposed T2DRL outperforms the other algorithms. Its performance is 8.28\% better than the DDPG-based T2DRL, 22.06\% better than SCHRS, and 24.43\% better than RCARS when the edge server's caching space is 20 GB; it is 14.01\% better than the DDPG-based T2DRL, 31.09\% better than SCHRS, and 57.56\% better than RCARS when the caching space is 32 GB.

Fig.~\ref{fig:Objective_value_different_caching} shows the effect of the edge server's caching capacity on the total utility~\eqref{eq:problem1}. We observe that as the edge server's caching space increases, the total utility~\eqref{eq:problem1} decreases. This is because the default preference weight factor $\alpha$ is set to 0.7, placing greater emphasis on AIGC service provisioning delay. As a result, a larger cache allows the edge server to store more GenAI models, enabling user requests to be fulfilled directly from the edge, which reduces both backhaul transmission time and image generation time. Overall, our proposed T2DRL outperforms the other algorithms. Its performance is 8.98\% better than the DDPG-based T2DRL, 22.31\% better than SCHRS, and 29.95\% better than RCARS when the edge server's caching space is 20 GB; it is 10.24\% better than the DDPG-based T2DRL, 27.87\% better than SCHRS, and 39.51\% better than RCARS when the caching space is 32 GB.


\section{Conclusion and Future Works} \label{sec:conclusion}
In this paper, we have addressed the problem of joint model caching and resource allocation in generative AI-enabled dynamic wireless networks. Our objective is to balance the trade-off between AIGC quality and service provisioning latency. To achieve this, we have first decomposed the problem into two subproblems: model caching on a long-timescale and resource allocation on a short-timescale. We have then leveraged a DDQN algorithm to tackle the model caching subproblem and proposed a D3PG algorithm to solve the resource allocation subproblem. Notably, the proposed D3PG algorithm make an innovative use of diffusion models in its actor network. Future research could explore the joint model caching and resource allocation problem for other types of AIGC services, such as text (e.g., \emph{ChatGPT}) and video (e.g., \emph{Sora}), as well as the potential for cooperation (e.g., revenue sharing) and competition among different edge servers in AIGC service provisioning.
\vspace{-3mm}


\bibliographystyle{IEEEtran}
\input{Main.bbl}

\vspace{-14mm}
\clearpage

\end{document}